\documentclass{article}

\PassOptionsToPackage{numbers, compress}{natbib}
\usepackage[final]{neurips_data_2023}

\usepackage[utf8]{inputenc} 
\usepackage[T1]{fontenc}    
\usepackage{hyperref}       
\usepackage{url}            
\usepackage{booktabs}       
\usepackage{amsfonts}       
\usepackage{nicefrac}       
\usepackage{microtype}      
\usepackage{xcolor}         
\usepackage{qiangstyle}
\usepackage{amsmath}
\usepackage{amsfonts}
\usepackage{hyperref}
\usepackage{amsmath}
\usepackage{amssymb}
\usepackage{mathtools}
\usepackage{amsthm}
\usepackage{hyperref}
\usepackage{url}
\usepackage{xcolor}
\usepackage{amssymb}
\usepackage{multirow}
\usepackage{bbding}
\usepackage{bbm}
\usepackage{listings}
\usepackage[T1]{fontenc}
\usepackage{booktabs, tabularx, colortbl} 
\usepackage{caption}
\usepackage{wrapfig,lipsum}
\usepackage{hhline}
\usepackage{tcolorbox}
\usepackage{graphicx}
\usepackage{bmpsize}

\definecolor{myRed}{rgb}{0.9, 0.5, 0.6}
\definecolor{myBlue}{rgb}{0.55, 0.70, 0.90}
\definecolor{myGreen}{rgb}{0.64, 0.83, 0.35}
\definecolor{myYellow}{rgb}{0.9, 0.75, 0.4}

\newcommand{\loosepar}{\looseness=-1}

\newcommand{\lldm}{\textsc{LLDM}}
\newcommand{\lb}{\textsc{LIBERO}}
\newcommand{\liberohundred}{\textsc{LIBERO-100}}
\newcommand{\liberolong}{\textsc{LIBERO-Long}}
\newcommand{\liberoninety}{\textsc{LIBERO-90}}
\newcommand{\liberospatial}{\textsc{LIBERO-Spatial}}
\newcommand{\liberoobject}{\textsc{LIBERO-Object}}
\newcommand{\liberox}{\textsc{LIBERO-X}}
\newcommand{\liberogoal}{\textsc{LIBERO-Goal}}
\newcommand{\challengeknowledge}{\textsc{(T1)}}
\newcommand{\challengealgorithm}{\textsc{(T3)}}
\newcommand{\challengearchitecture}{\textsc{(T2)}}
\newcommand{\challengeordering}{\textsc{(T4)}}
\newcommand{\challengepretraining}{\textsc{(T5)}}
\newcommand{\qone}{\textbf{Q1}}
\newcommand{\qtwo}{\textbf{Q2}}
\newcommand{\qthree}{\textbf{Q3}}
\newcommand{\qfour}{\textbf{Q4}}
\newcommand{\qfive}{\textbf{Q5}}
\newcommand{\qsix}{\textbf{Q6}}

\newcommand{\bcrnn}{\textsc{ResNet-RNN}}
\newcommand{\bct}{\textsc{ResNet-T}}
\newcommand{\bcvilt}{\textsc{ViT-T}}

\newcommand{\er}{\textsc{ER}}
\newcommand{\ewc}{\textsc{EWC}}
\newcommand{\packnet}{\textsc{PackNet}}
\newcommand{\seql}{\textsc{SeqL}}
\newcommand{\mtl}{\textsc{MTL}}

\newcommand{\robomimic}{\texttt{Robomimic}}
\newcommand{\robosuite}{\texttt{Robosuite}}

\newcommand{\myparagraph}[1]{\textbf{#1}~~~}
\newcommand{\mysection}[1]{\section{#1}}
\newcommand{\mysubsection}[1]{\subsection{#1}}

\newcommand{\fs}[1]{\footnotesize $\pm$#1}

\usepackage[capitalize,noabbrev]{cleveref}
\definecolor{top1_boxit_color}{RGB}{186,51,121}
\definecolor{top2_boxit_color}{RGB}{186,51,121}
\definecolor{top3_boxit_color}{RGB}{186,51,121}
\newcommand{\toponeboxit}{\cellcolor{top1_boxit_color!90}}
\newcommand{\toptwoboxit}{\cellcolor{top2_boxit_color!50}}
\newcommand{\topthreeboxit}{\cellcolor{top3_boxit_color!20}}

\title{LIBERO: Benchmarking Knowledge Transfer for Lifelong Robot Learning}

\author{%
  $^\dagger$Bo Liu\thanks{Equal contribution.}, $^\dagger$Yifeng Zhu$^*$, $^\ddagger$Chongkai Gao$^*$, $^{\dagger}$Yihao Feng\\\textbf{$^\dagger$Qiang Liu, $^{\dagger}$Yuke Zhu, $^{\dagger,\mathsection}$Peter Stone} \\
  $^\dagger$The University of Texas at Austin, $^{\mathsection}$Sony AI, $^\ddagger$Tsinghua University\\
  \texttt{\{bliu,yifengz,lqiang,yukez,pstone\}@cs.utexas.edu}\\
  \texttt{yihao.ac@gmail.com, gck20@mails.tsinghua.edu.cn}
}

\begin{document}

\maketitle

\begin{abstract}
Lifelong learning offers a promising paradigm of building a generalist agent that learns and adapts over its lifespan. 
Unlike traditional lifelong learning problems in image and text domains, which primarily involve the transfer of declarative knowledge of entities and concepts, lifelong learning in decision-making (\lldm{}) also necessitates the transfer of procedural knowledge, such as actions and behaviors.
To advance research in \lldm{}, we introduce \lb{}, a novel benchmark of lifelong learning for robot manipulation. Specifically, \lb{} highlights five key research topics in \lldm{}: \textbf{1)} how to efficiently transfer declarative knowledge, procedural knowledge, or the mixture of both; \textbf{2)} how to design effective policy architectures and \textbf{3)} effective algorithms for \lldm{}; \textbf{4)} the robustness of a lifelong learner with respect to task ordering; and \textbf{5)} the effect of model pretraining for \lldm{}.
We develop an extendible \emph{procedural generation} pipeline that can in principle generate infinitely many tasks. For benchmarking purpose, we create four task suites (130 tasks in total) that we use to investigate the above-mentioned research topics. To support sample-efficient learning, we provide high-quality human-teleoperated demonstration data for all tasks.
Our extensive experiments present several insightful or even \emph{unexpected} discoveries: sequential finetuning outperforms existing lifelong learning methods in forward transfer, no single visual encoder architecture excels at all types of knowledge transfer, and naive supervised pretraining can hinder agents' performance in the subsequent \lldm{}.\footnote{Check the website at \url{https://libero-project.github.io} for the code and the datasets.}
\end{abstract}
\mysection{Introduction}
A longstanding goal in machine learning is to develop a generalist agent that can perform a wide range of tasks. While multitask learning~\citep{caruana1997multitask} is one approach, it is computationally demanding and not adaptable to ongoing changes. Lifelong learning~\citep{thrun1995lifelong}, however, offers a practical solution by amortizing the learning process over the agent's lifespan. Its goal is to leverage prior knowledge to facilitate learning new tasks (forward transfer) and use the newly acquired knowledge to enhance performance on prior tasks (backward transfer).

\begin{figure*}[ht!]
    \centering
    \includegraphics[width=\textwidth]{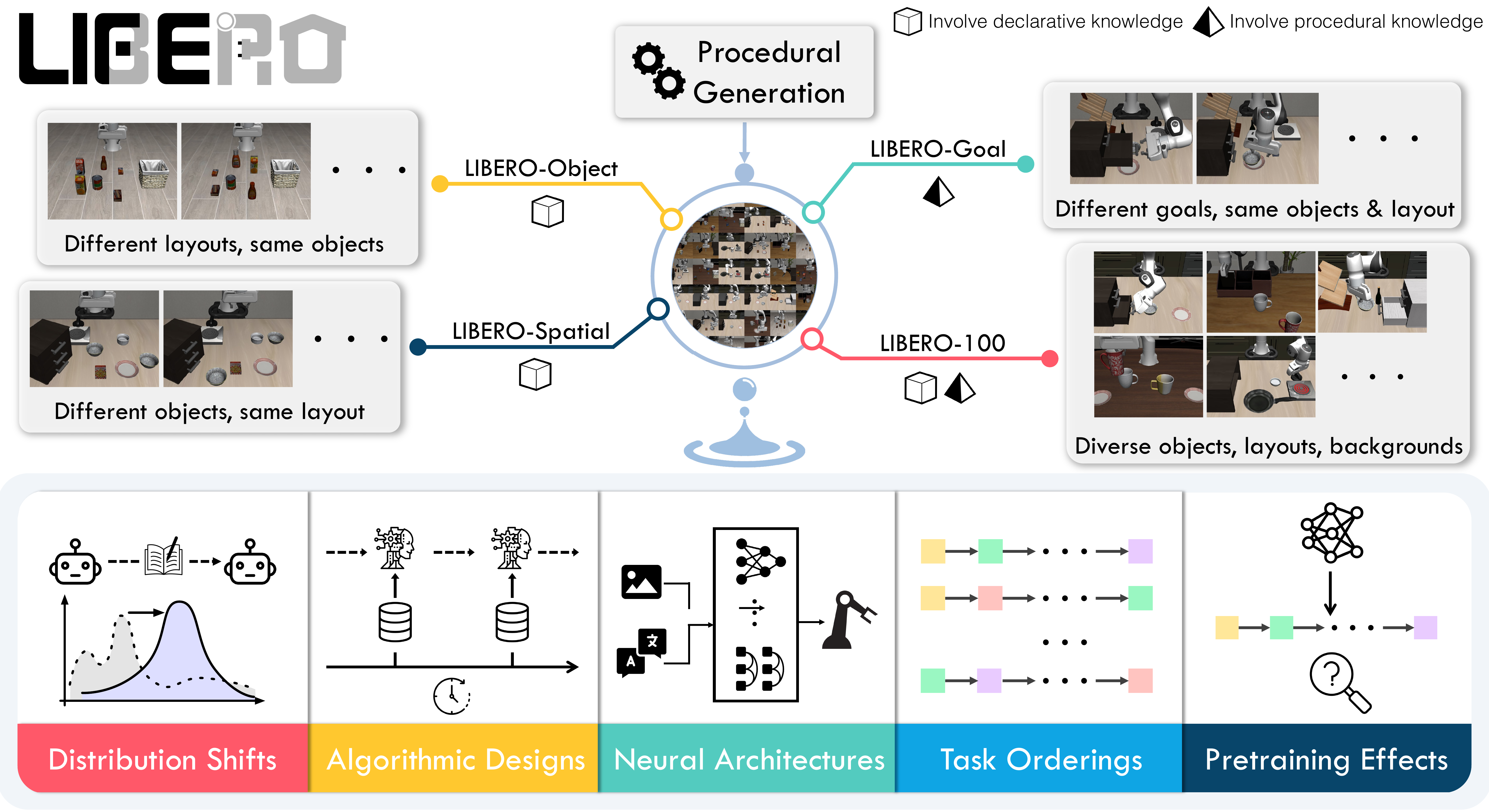}
    \caption{
    \textbf{Top}: \lb{} has four procedurally-generated task suites: \liberospatial{}, \liberoobject{}, and \liberogoal{} have 10 tasks each and require transferring knowledge about spatial relationships, objects, and task goals; \liberohundred{} has 100 tasks and requires the transfer of entangled knowledge.
    \textbf{Bottom}: we investigate five key research topics in \lldm{} on \lb{}.
    }
    \label{fig:libero-overview}
\end{figure*}

The main body of the lifelong learning literature has focused on how agents transfer \emph{declarative} knowledge in visual or language tasks, which pertains to \emph{declarative knowledge} about entities and concepts~\citep{biesialska2020continual, mai2022online}. Yet it is understudied how agents transfer knowledge in decision-making tasks, which involves a mixture of both \emph{declarative} and \emph{procedural} knowledge (knowledge about how to \emph{do} something). Consider a scenario where a robot, initially trained to retrieve juice from a fridge, fails after learning new tasks. This could be due to forgetting the juice or fridge's location (declarative knowledge) or how to open the fridge or grasp the juice (procedural knowledge). So far, we lack methods to systematically and quantitatively analyze this complex knowledge transfer.

To bridge this research gap, this paper introduces a new simulation benchmark, LIfelong learning BEchmark on RObot manipulation tasks, \lb{}, to facilitate the systematic study of lifelong learning in decision making (\lldm{}). An ideal \lldm{} testbed should enable continuous learning across an expanding set of diverse tasks that share concepts and actions. \lb{} supports this through a procedural generation pipeline for endless task creation, based on robot manipulation tasks with shared visual concepts (declarative knowledge) and interactions (procedural knowledge).

For benchmarking purpose, \lb{} generates 130 language-conditioned robot manipulation tasks inspired by human activities~\citep{grauman2022ego4d} and, grouped into four suites. The four task suites are designed to examine distribution shifts in the object types, the spatial arrangement of objects, the task goals, or the mixture of the previous three (top row of Figure~~\ref{fig:libero-overview}). \lb{} is scalable, extendable, and designed explicitly for studying lifelong learning in robot manipulation. To support efficient learning, we provide high-quality, human-teleoperated demonstration data for all 130 tasks. 

We present an initial study using \lb{} to investigate five major research topics in \lldm{} (Figure~\ref{fig:libero-overview}): \textbf{1)} knowledge transfer with different types of distribution shift; \textbf{2)} neural architecture design;  \textbf{3)}  lifelong learning algorithm design; \textbf{4)} robustness of the learner to task ordering; and \textbf{5)} how to leverage pre-trained models in \lldm{} (bottom row of Figure~\ref{fig:libero-overview}). We perform extensive experiments across different policy architectures and different lifelong learning algorithms. Based on our experiments, we make several insightful or even \textbf{unexpected} observations:
\begin{enumerate}
    \item Policy architecture design is as crucial as lifelong learning algorithms. The transformer architecture is better at abstracting temporal information than a recurrent neural network. Vision transformers work well on tasks with rich visual information (e.g., a variety of objects). Convolution networks work well when tasks primarily need procedural knowledge.
    \item While the lifelong learning algorithms we evaluated are effective at preventing forgetting, they generally perform \emph{worse} than sequential finetuning in terms of forward transfer.
    \item Our experiment shows that using pretrained language embeddings of semantically-rich task descriptions yields performance \emph{no better} than using those of the task IDs.
    \item Basic supervised pretraining on a large-scale offline dataset can have a \emph{negative} impact on the learner's downstream performance in \lldm{}.
\end{enumerate}

\mysection{Background}
This section introduces the problem formulation and defines key terms used throughout the paper.

\mysubsection{Markov Decision Process for Robot Learning}
A robot learning problem can be formulated as a finite-horizon Markov Decision Process:
$
\mathcal{M} = (\mathcal{S}, \mathcal{A}, \mathcal{T}, H, ~{\color{purple}{\mu_0, R}}).
$
Here, $\mathcal{S}$ and $\mathcal{A}$ are the state and action spaces of the robot. $\mu_0$ is the initial state distribution, $R: \mathcal{S}\times \mathcal{A} \rightarrow \mathbb{R}$ is the reward function, and $\mathcal{T}: \mathcal{S} \times \mathcal{A} \rightarrow \mathcal{S}$ is the transition function. In this work, we assume a sparse-reward setting and replace $R$ with a goal predicate $g: \mathcal{S} \rightarrow \{0, 1\}$. The robot's objective is to learn a policy $\pi$ that maximizes the expected return:
$
\max_\pi J(\pi) = \mathbb{E}_{s_t, a_t \sim \pi, \mu_0} [\sum_{t=1}^{H} g(s_t)].
$

\mysubsection{Lifelong Robot Learning Problem}
In a \emph{lifelong robot learning problem}, a robot sequentially learns over $K$ tasks $\{T^1, \dots, T^K\}$ with a single policy $\pi$. We assume $\pi$ is conditioned on the task, i.e., $\pi(\cdot \mid s; T)$. For each task, $T^k \equiv (\mu_0^k, g^k)$ is defined by the initial state distribution $\mu_0^k$ and the goal predicate $g^k$.\footnote{Throughout the paper, a superscript/subscript is used to index the task/time step.} We assume $\mathcal{S}, \mathcal{A}, \mathcal{T}, H$ are the same for all tasks. Up to the $k$-th task $T^k$, the robot aims to optimize
\begin{equation}
    \max_\pi~ J_{\text{LRL}}(\pi) = \frac{1}{k}\sum_{p=1}^k \bigg[ \mathop{\mathbb{E}}\limits_{s^p_t, a^p_t \sim \pi(\cdot;T^p),~ \mu_0^p} \bigg[\sum_{t=1}^L g^p(s_t^p) \bigg] \bigg].
    \label{eq:LRL}
\end{equation}
An important feature of the lifelong setting is that the agent loses access to the previous $k-1$ tasks when it learns on task $T^k$.

\myparagraph{Lifelong Imitation Learning} Due to the challenge of sparse-reward reinforcement learning,
 we consider a practical alternative setting where a user would provide a small demonstration dataset for each task in the sequence. Denote $D^k = \{ \tau^k_i \}_{i=1}^N$ as $N$ demonstrations for task $T^k$. Each $\tau^k_i = (o_0, a_0, o_1, a_1, \dots, o_{l^k})$ where $l^k \leq H$. Here, $o_t$ is the robot's sensory input, including the perceptual observation and the information about the robot's joints and gripper. In practice, the observation $o_t$ is often non-Markovian. Therefore, following works in partially observable MDPs~\citep{hausknecht2015deep}, we represent $s_t$ by the aggregated history of observations, i.e. $s_t \equiv o_{\leq t} \triangleq (o_0, o_1, \dots, o_t) $.
This results in the \emph{lifelong imitation learning problem} with the same objective as in Eq.~\eqref{eq:LRL}. But during training, we perform behavioral cloning~\citep{bain1995framework} with the following surrogate objective function:
\begin{equation}
 \min_\pi~ J_{\text{BC}}(\pi) = \frac{1}{k}\sum_{p=1}^k \mathop{\mathbb{E}}\limits_{o_t, a_t \sim D^p} \bigg[ \sum_{t=0}^{l^p} \mathcal{L}\big(\pi(o_{\leq t}; T^p), a^p_t\big)\bigg]\,,
    \label{eq:bc}
\end{equation}
where $\mathcal{L}$ is a supervised learning loss, e.g., the negative log-likelihood loss, and $\pi$ is a Gaussian mixture model. Similarly, we assume $\{D^p: p < k\}$ are not fully available when learning $T^k$.
\mysection{Research Topics in \lldm{}}
\label{sec:research-challenge}
We outline five major research topics in \lldm{} that motivate the design of \lb{} and our study.

\myparagraph{\challengeknowledge{} Transfer of Different Types of Knowledge} 
In order to accomplish a task such as \emph{put the ketchup next to the plate in the basket}, a robot must understand the concept \emph{ketchup}, the location of the \emph{plate/basket}, and how to \emph{put} the ketchup in the basket. 
Indeed, robot manipulation tasks in general necessitate different types of knowledge, making it hard to determine the cause of failure. We present four task suites in Section~\ref{sec:libero-suite}: three task suites for studying the transfer of knowledge about spatial relationships, object concepts, and task goals in a disentangled manner, and one suite for studying the transfer of mixed types of knowledge.

\myparagraph{\challengearchitecture{} Neural Architecture Design} 
An important research question in \lldm{} is how to design effective neural architectures to abstract the multi-modal observations (images, language descriptions, and robot states) and transfer only relevant knowledge when learning new tasks.

\myparagraph{\challengealgorithm{} Lifelong Learning Algorithm Design} Given a policy architecture, it is crucial to determine what learning algorithms to apply for \lldm{}. Specifically, the sequential nature of \lldm{} suggests that even minor forgetting over successive steps can potentially lead to a total failure in execution. As such, we consider the design of lifelong learning algorithms to be an open area of research in \lldm{}.

\myparagraph{\challengeordering{} Robustness to Task Ordering}
It is well-known that task curriculum influences policy learning \cite{bengio2009curriculum,narvekar2020curriculum}. A robot in the real world, however, often cannot choose which task to encounter first. Therefore, a good lifelong learning algorithm should be robust to different task orderings.

\myparagraph{\challengepretraining{} Usage of Pretrained Models} In practice, robots will be most likely pretrained on large datasets in factories before deployment~\citep{kaelbling2020foundation}. However, it is not well-understood whether or how pretraining could benefit subsequent \lldm{}. 
\mysection{LIBERO}
This section introduces the components in \lb{}: the procedural generation pipeline that allows the never-ending creation of tasks (Section~\ref{sec:procedural}), the four task suites we generate for benchmarking (Section~\ref{sec:libero-suite}), five algorithms (Section~\ref{sec:algo}), and three neural architectures (Section~\ref{sec:arch}). 
\begin{figure*}[t!]
    \centering
    \includegraphics[width=\textwidth]{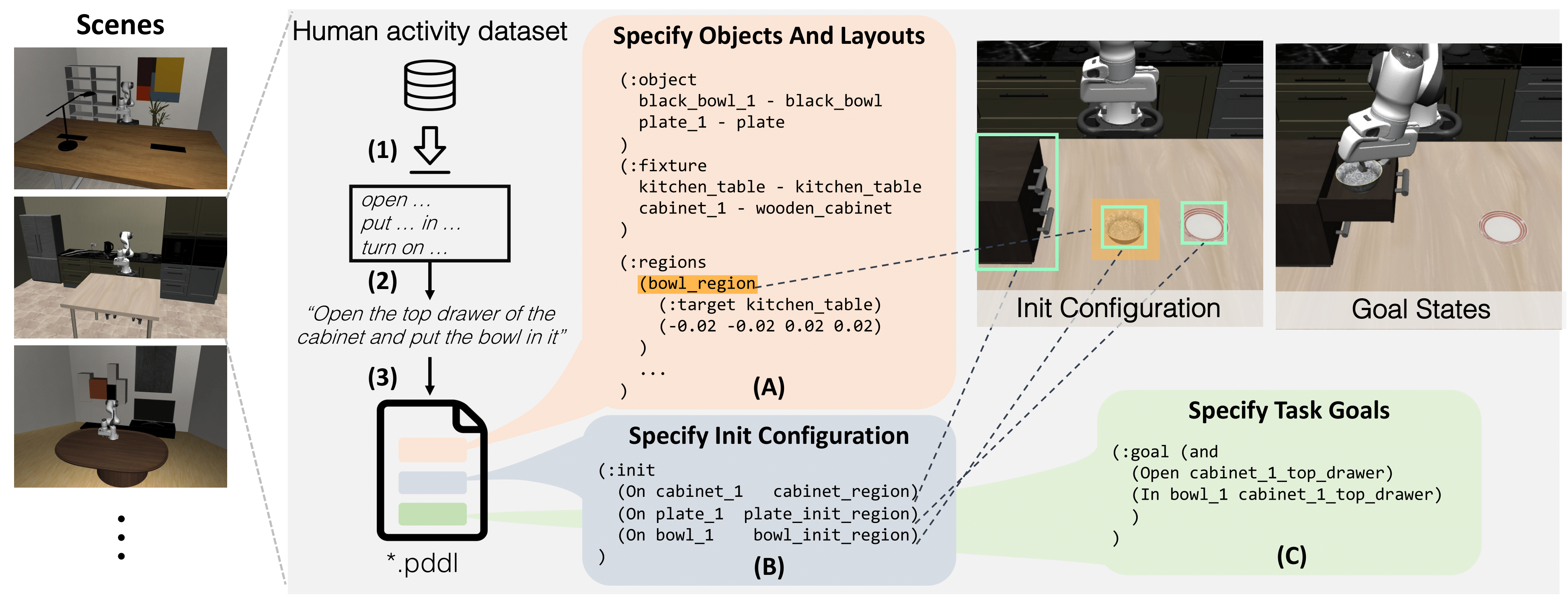}

    \caption{\lb{}'s procedural generation pipeline:  Extracting behavioral templates from a large-scale human activity dataset \textbf{(1)}, Ego4D, for generating task instructions \textbf{(2)}; Based on the task description, selecting the scene and generating the PDDL description file \textbf{(3)} that specifies the objects and layouts \textbf{(A)}, the initial object configurations \textbf{(B)}, and the task goal \textbf{(C)}.
    }
    \label{fig:libero-procedural-generation}
\end{figure*}

\mysubsection{Procedural Generation of Tasks}
\label{sec:procedural}
Research in \lldm{} requires a systematic way to create new tasks while maintaining task diversity and relevance to existing tasks. \lb{} procedurally generates new tasks in three steps: \textbf{1)} extract behavioral templates from language annotations of human activities and generate sampled tasks described in natural language based on such templates; \textbf{2)} specify an initial object distribution given a task description; and \textbf{3)} specify task goals using a propositional formula that aligns with the language instructions.
Our generation pipeline is built on top of \robosuite{}~\citep{zhu2020robosuite}, a modular manipulation simulator that offers seamless integration. Figure~\ref{fig:libero-procedural-generation} illustrates an example of task creation using this pipeline, and each component is expanded upon below.

\myparagraph{Behavioral Templates and Instruction Generation} Human activities serve as a fertile source of tasks that can inspire and generate a vast number of manipulation tasks. 
We choose a large-scale activity dataset, Ego4D~\citep{grauman2022ego4d}, which includes a large variety of everyday activities with language annotations. We pre-process the dataset by extracting the language descriptions and then summarize them into a large set of commonly used language templates. After this pre-processing step, we use the templates and select objects available in the simulator to generate a set of task descriptions in the form of language instructions. For example, we can generate an instruction ``Open the drawer of the cabinet'' from the template ``Open ...''.

\myparagraph{Initial State Distribution ($\mu_0$)} 
To specify $\mu_0$, we first sample a scene layout that matches the objects/behaviors in a provided instruction. For instance, a kitchen scene is selected for an instruction \textit{Open the top drawer of the cabinet and put the bowl in it}. Then, the details about $\mu_0$ are generated in the PDDL language~\citep{mcdermott1998pddl,srivastava2022behavior}.
Concretely, $\mu_0$ contains information about object categories and their placement (Figure~\ref{fig:libero-procedural-generation}-\textbf{(A)}), and their initial status (Figure~\ref{fig:libero-procedural-generation}-\textbf{(B)}).

\myparagraph{Goal Specifications $(g)$} Based on $\mu_{0}$ and the language instruction, we specify the task goal using a conjunction of predicates. Predicates include \emph{unary predicates} that describe the properties of an object, such as \texttt{Open}(X) or \texttt{TurnOff}(X), and \emph{binary predicates} that describe spatial relations between objects, such as \texttt{On}(A, B) or \texttt{In}(A, B). An example of the goal specification using PDDL language can be found in Figure~\ref{fig:libero-procedural-generation}-\textbf{(C)}. The simulation terminates when all predicates are verified true.

\mysubsection{Task Suites}
\label{sec:libero-suite}
While the pipeline in Section~\ref{sec:procedural} supports the generation of an unlimited number of tasks, we offer fixed sets of tasks for benchmarking purposes.
\lb{} has four task suites: \liberospatial, \liberoobject, \liberogoal, and \liberohundred. The first three task suites are curated to disentangle the transfer of \emph{declarative} and \emph{procedural} knowledge (as mentioned in~\challengeknowledge{}), while \liberohundred{} is a suite of 100 tasks with entangled knowledge transfer. 

\myparagraph{\liberox{}} \liberospatial{}, \liberoobject{}, and \liberogoal{} all have 10 tasks\footnote{
A suite of 10 tasks is enough to observe catastrophic forgetting while maintaining computation efficiency.
} 
and are designed to investigate the controlled transfer of knowledge about spatial information (declarative), objects (declarative), and task goals (procedural).
Specifically, all tasks in \liberospatial{} request the robot to place a bowl, among the same set of objects, on a plate. But there are two identical bowls that differ only in their location or spatial relationship to other objects. Hence, to successfully complete \liberospatial{}, the robot needs to continually learn and memorize new spatial relationships.
All tasks in \liberoobject{} request the robot to pick-place a unique object. 
Hence, to accomplish \liberoobject{}, the robot needs to continually learn and memorize new object types.
All tasks in \liberogoal{} share the same objects with fixed spatial relationships but differ only in the task goal. Hence, to accomplish \liberogoal{}, the robot needs to continually learn new knowledge about motions and behaviors.
More details are in Appendix~\ref{appendix:task}.

\myparagraph{\liberohundred}  \liberohundred{} contains 100 tasks that entail diverse object interactions and versatile motor skills. 
In this paper, we split \liberohundred{} into 90 short-horizon tasks (\liberoninety{}) and 10 long-horizon tasks (\liberolong{}). \liberoninety{} serves as the data source for pretraining~\textbf{\challengepretraining{}} and \liberolong{} for downstream evaluation of lifelong learning algorithms.

\mysubsection{Lifelong Learning Algorithms}
\label{sec:algo}
We implement three representative lifelong learning algorithms to facilitate research in algorithmic design for \lldm{}. Specifically, we implement Experience Replay (\er{})~\citep{chaudhry2019tiny}, Elastic Weight Consolidation (\ewc{})~\citep{kirkpatrick2017overcoming}, and~\packnet{}~\citep{mallya2018packnet}. We pick \er{}, \ewc{}, and \packnet{} because they correspond to the memory-based, regularization-based, and dynamic-architecture-based methods for lifelong learning. In addition, prior research~\cite{Woczyk2021ContinualWA} has discovered that they are state-of-the-art methods. Besides these three methods, we also implement sequential finetuning (\seql{}) and multitask learning (\mtl{}), which serve as a lower bound and upper bound for lifelong learning algorithms, respectively. More details about the algorithms are in Appendix~\ref{appendix:llalgo}.

\mysubsection{Neural Network Architectures} 
\label{sec:arch}

\label{sec:method-architecture}
We implement three vision-language policy networks, \bcrnn{}, \bct{}, and \bcvilt{}, that integrate visual, temporal, and linguistic information for \lldm{}.
Language instructions of tasks are encoded using pretrained BERT embeddings~\cite{devlin2018bert}. 
The \bcrnn{}~\citep{mandlekar2021matters} uses a ResNet as the visual backbone that encodes per-step visual observations and an LSTM as the temporal backbone to process a sequence of encoded visual information. The language instruction is incorporated into the ResNet features using the FiLM method~\citep{perez2018film} and added to the LSTM inputs, respectively.
\bct{} architecture~\citep{zhu2022viola} uses a similar ResNet-based visual backbone, but a transformer decoder~\citep{vaswani2017attention} as the temporal backbone to process outputs from ResNet, which are a temporal sequence of visual tokens. The language embedding is treated as a separate token in inputs to the transformer alongside the visual tokens.
The \bcvilt{} architecture~\citep{kim2021vilt}, which is widely used in visual-language tasks, uses a Vision Transformer (ViT) as the visual backbone and a transformer decoder as the temporal backbone. The language embedding is treated as a separate token in inputs of both ViT and the transformer decoder. All the temporal backbones output a latent vector for every decision-making step. We compute the multi-modal distribution over manipulation actions using a Gaussian-Mixture-Model (GMM) based output head~\citep{bishop1994mixture, mandlekar2021matters, wang2023mimicplay}. In the end, a robot executes a policy by sampling a continuous value for end-effector action from the output distribution. Figure~\ref{fig:architectures} visualizes the three architectures.\loosepar{}

For all the lifelong learning algorithms and neural architectures, we use behavioral cloning (BC)~\citep{bain1995framework} to train policies for individual tasks (See \eqref{eq:bc}). BC allows for efficient policy learning such that we can study lifelong learning algorithms with limited computational resources. 
To train BC, we provide 50 trajectories of high-quality demonstrations for every single task in the generated task suites. The demonstrations are collected by human experts through teleoperation with 3Dconnexion Spacemouse.

\mysection{Experiments}
Experiments are conducted as an initial study for the five research topics mentioned in Section~\ref{sec:research-challenge}. We first introduce the evaluation metric used in experiments, and present analysis of empirical results in \lb{}. The detailed experimental setup is in Appendix~\ref{appendix:exp-setting}. Our experiments focus on addressing the following research questions:


\qone{}: How do different architectures/LL algorithms perform under specific distribution shifts?\\
\qtwo{}: To what extent does neural architecture impact knowledge transfer in \lldm{}, and are there any discernible patterns in the specialized capabilities of each architecture? \\
\qthree{}: How do existing algorithms from lifelong supervised learning perform on \lldm{} tasks? \\
\qfour{}: To what extent does language embedding affect knowledge transfer in \lldm{}? \\
\qfive{}: How robust are different LL algorithms to task ordering in \lldm{}? \\
\qsix{}: Can supervised pretraining improve downstream lifelong learning performance in \lldm{}?

\begin{figure}[h!]
    \centering
    \includegraphics[width=0.75\textwidth]{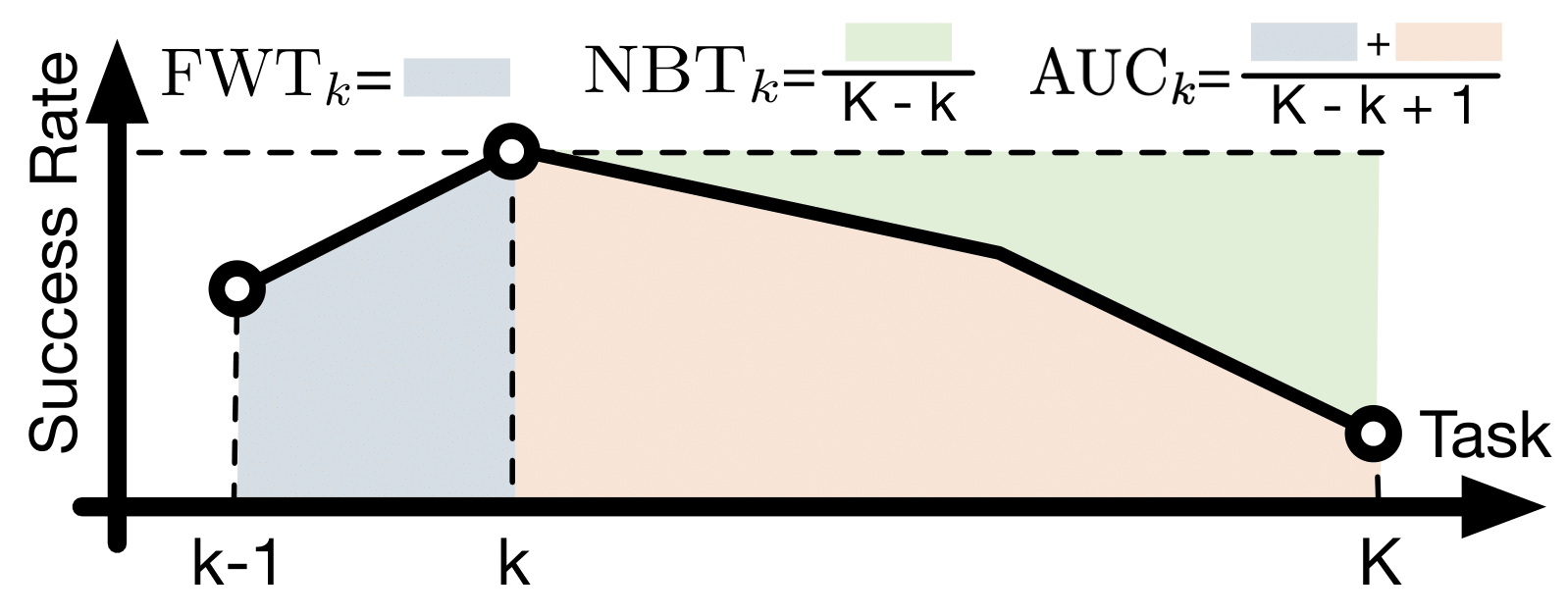}
    \caption{Metrics for \lldm{}.}
    \label{fig:metrics}
\end{figure}

\mysubsection{Evaluation Metrics}
We report three metrics: FWT (forward transfer)~\citep{diaz2018don}, NBT (negative backward transfer), and AUC (area under the success rate curve). All metrics are computed in terms of success rate, as previous literature has shown that the success rate is a more reliable metric than training loss for manipulation policies~\citep{mandlekar2021matters} (Detailed explanation in Appendix~\ref{appendix:loss-success-rates}). Lower NBT means a policy has better performance in the previously seen tasks, higher FWT means a policy learns faster on a new task, and higher AUC means an overall better performance considering both NBT and FWT. Specifically, denote $c_{i,j,e}$ as the agent's success rate on task $j$ when it learned over $i-1$ previous tasks and has just learned $e$ epochs ($e \in \{0,5,\dots,50\}$) on task $i$. Let $c_{i,i}$ be the best success rate over all evaluated epochs $e$ for the current task $i$ (i.e., $c_{i,i} = \max_e c_{i,i,e}$). Then, we find the earliest epoch $e^*_i$ in which the agent achieves the best performance on task $i$ (i.e., $e^*_i = \argmin_e c_{i,i,e_i} = c_{i,i}$), and assume for all $e \geq e^*_i$, $c_{i,i,e} = c_{i,i}$.\footnote{In practice, it's possible that the agent's performance on task $i$ is not monotonically increasing due to the variance of learning. But we keep the best checkpoint among those saved at epochs $\{e\}$ as if the agent stops learning after $e^*_i$.}
Given a different task $j \neq i$, we define $c_{i,j} = c_{i,j,e^*_i}$. Then the three metrics are defined:
\begin{equation}
\begin{split}
&\text{FWT} = \sum_{k\in[K]} \frac{\text{FWT}_k}{K},~~~\text{FWT}_k = \frac{1}{11}\sum_{e \in \{0 \dots 50\}} c_{k,k,e} \\
&\text{NBT} = \sum_{k\in[K]} \frac{\text{NBT}_k}{K},~~~~\text{NBT}_k =  \frac{1}{K-k} \sum_{\tau = k+1}^K \big(c_{k, k} - c_{\tau, k}\big) \\
&\text{AUC} = \sum_{k\in[K]} \frac{\text{AUC}_k}{K},~~~\text{AUC}_k =\frac{1}{K-k+1} \big(\text{FWT}_k + \sum_{\tau=k+1}^K c_{\tau, k}\big)\\
\end{split}
\end{equation}
A visualization of these metrics is provided in Figure~\ref{fig:metrics}. 

\mysubsection{Experimental Results}
\label{experiment}
We present empirical results to address the research questions. Please refer to Appendix~\ref{appendix:additional-result} for the full results across all algorithms, policy architectures, and task suites.

\myparagraph{Study on the Policy's Neural Architectures (\qone{}, \qtwo{}) } Table~\ref{tab:architecture} reports the agent's lifelong learning performance using the three different neural architectures on the four task suites. Results are reported when \er{} and \packnet{} are used as they demonstrate the best lifelong learning performance across all task suites.
\begin{table}[!ht]
    \centering
    \resizebox{\textwidth}{!}{%
    \begin{tabular}{ l  c c c c c c}
    \toprule 
    \multirow{ 2}{*}{Policy Arch.} & \multicolumn{3}{c}{\er{}} & \multicolumn{3}{c}{\packnet{}} \\
    \cmidrule(lr){2-4} \cmidrule(lr){5-7}
    & FWT($\uparrow$) & NBT($\downarrow$) & AUC($\uparrow$) & FWT($\uparrow$) & NBT($\downarrow$) & AUC($\uparrow$) \\
    \midrule
    & \multicolumn{6}{c}{\liberolong{}} \\
    \cmidrule(lr){2-7}
\bcrnn                    &  0.16 \fs {   0.02 } &  \textbf{0.16}  \fs {   0.02 } &  0.08  \fs {  0.01 } &  0.13 \fs {   0.00 } &  0.21  \fs {   0.01 } &  0.03  \fs {  0.00 }\\
\bct                      &  \textbf{0.48} \fs {   0.02 } &  0.32  \fs {   0.04 } &  \textcolor{purple}{\textbf{0.32}}  \fs {  0.01 } &  0.22 \fs {   0.01 } &   \textcolor{purple}{\textbf{0.08}}  \fs {   0.01 } &  0.25  \fs {  0.00 }\\
\bcvilt                   &  0.38 \fs {   0.05 } &  0.29  \fs {   0.06 } &  0.25  \fs {  0.02 } &   \textcolor{purple}{\textbf{0.36}} \fs {   0.01 } &  0.14  \fs {   0.01 } &   \textcolor{purple}{\textbf{0.34}}  \fs {  0.01 }\\
    \midrule 
        & \multicolumn{6}{c}{\liberospatial{}} \\
    \cmidrule(lr){2-7}
\bcrnn                    &  0.40 \fs {   0.02 } &  0.29  \fs {   0.02 } &  0.29  \fs {  0.01 } &  0.27 \fs {   0.03 } &  0.38  \fs {   0.03 } &  0.06  \fs {  0.01 }\\
\bct                      &  \textbf{0.65} \fs {   0.03 } &  \textbf{0.27}  \fs {   0.03 } &  \textbf{0.56}  \fs {  0.01 } &  0.55 \fs {   0.01 } &  \textbf{0.07}  \fs {   0.02 } &  \textbf{0.63}  \fs {  0.00 }\\
\bcvilt                   &  0.63 \fs {   0.01 } &  0.29  \fs {   0.02 } &  0.50  \fs {  0.02 } &  \textbf{0.57} \fs {   0.04 } &  0.15  \fs {   0.00 } &  0.59  \fs {  0.03 }\\
    \midrule 
        & \multicolumn{6}{c}{\liberoobject{}} \\
    \cmidrule(lr){2-7}
\bcrnn                    &  0.30 \fs {   0.01 } &  \textbf{0.27}  \fs {   0.05 } &  0.17  \fs {  0.05 } &  0.29 \fs {   0.02 } &  0.35  \fs {   0.02 } &  0.13  \fs {  0.01 }\\
\bct                      &  0.67 \fs {   0.07 } &  0.43  \fs {   0.04 } &  0.44  \fs {  0.06 } &  \textbf{0.60} \fs {   0.07 } &  \textbf{0.17}  \fs {   0.05 } &  \textbf{0.60}  \fs {  0.05 }\\
\bcvilt                   & \textbf{0.70} \fs {   0.02 } &  0.28  \fs {   0.01 } &  \textbf{0.57}  \fs {  0.01 } &  0.58 \fs {   0.03 } &  0.18  \fs {   0.02 } &  0.56  \fs {  0.04 }\\
    \midrule
        & \multicolumn{6}{c}{\liberogoal{}} \\
    \cmidrule(lr){2-7}
\bcrnn                    &  0.41 \fs {   0.00 } &  0.35  \fs {   0.01 } &  0.26  \fs {  0.01 } &  0.32 \fs {   0.03 } &  0.37  \fs {   0.04 } &  0.11  \fs {  0.01 }\\
\bct                      &   \textcolor{purple}{\textbf{0.64}} \fs {   0.01 } &  \textbf{0.34}  \fs {   0.02 } &   \textcolor{purple}{\textbf{0.49}}  \fs {  0.02 } &  0.63 \fs {   0.02 } &  \textbf{0.06}  \fs {   0.01 } &  0.75  \fs {  0.01 }\\
\bcvilt                   &  0.57 \fs {   0.00 } &  0.40  \fs {   0.02 } &  0.38  \fs {  0.01 } &  \textbf{0.69} \fs {   0.02 } &  0.08  \fs {   0.01 } &  \textbf{0.76}  \fs {  0.02 }\\
\bottomrule
\end{tabular}
}
\vspace{5pt}
\caption{Performance of the three neural architectures using \er{} and \packnet{} on the four task suites. Results are averaged over three seeds and we report the mean and standard error. The best performance is \textbf{bolded}, and colored in \textcolor{purple}{\textbf{purple}} if the improvement is statistically significant over other neural architectures, when a two-tailed, Student’s
t-test under equal sample sizes and unequal variance is applied with a $p$-value of 0.05.}
\label{tab:architecture}
\end{table}

\textcolor{purple}{\textit{Findings:}} First, we observe that \bct{} and \bcvilt{} work much better than \bcrnn{} on average, indicating that using a transformer on the ``temporal" level could be a better option than using an RNN model. Second, the performance difference among different architectures depends on the underlying lifelong learning algorithm. If \packnet{} (a dynamic architecture approach) is used, we observe no significant performance difference between \bct{} and \bcvilt{} except on the \liberolong{} task suite where \bcvilt{} performs much better than \bct{}. In contrast, if \er{} is used, we observe that \bct{} performs better than \bcvilt{} on all task suites except \liberoobject{}. This potentially indicates that the ViT architecture is better at processing visual information with more object varieties than the ResNet architecture when the network capacity is sufficiently large (See the \mtl{} results in Table~\ref{tab:algorithm} on \liberoobject{} as the supporting evidence). The above findings shed light on how one can improve architecture design for better processing of spatial and temporal information in \lldm{}.

\myparagraph{Study on Lifelong Learning Algorithms (\qone{}, \qthree{})} Table~\ref{tab:algorithm-short} reports the lifelong learning performance of the three lifelong learning algorithms, together with the \seql{} and \mtl{} baselines. All experiments use the same \bct{} architecture as it performs the best across all policy architectures.

\begin{table}[ht!]
    \centering
    \vspace{5pt}
    \resizebox{\textwidth}{!}{%
    \begin{tabular}{ l c c c c c c}
    \toprule
    Lifelong Algo. & FWT($\uparrow$) & NBT($\downarrow$) & AUC($\uparrow$) & FWT($\uparrow$) & NBT($\downarrow$) & AUC($\uparrow$) \\
    \midrule 
    & \multicolumn{3}{c}{\liberolong{}} & \multicolumn{3}{c}{\liberospatial{}} \\
    \cmidrule(lr){2-4} \cmidrule(l){5-7} 
 \seql                &  \textbf{0.54} \fs {   0.01 } &  0.63  \fs {   0.01 } &  0.15  \fs {  0.00 }  & \textbf{0.72} \fs {   0.01 } &  0.81  \fs {   0.01 } &  0.20  \fs {  0.01 } \\                                                                                                                                         
 \er                  &  0.48 \fs {   0.02 } &  0.32  \fs {   0.04 } &  \textcolor{purple}{\textbf{0.32}}  \fs {  0.01 } &  0.65 \fs {   0.03 } &  0.27  \fs {   0.03 } &  0.56  \fs {  0.01 } \\                                                                                                                                         
 \ewc                 &  0.13 \fs {   0.02 } &  0.22  \fs {   0.03 } &  0.02  \fs {  0.00 } &  0.23 \fs {   0.01 } &  0.33  \fs {   0.01 } &  0.06  \fs {  0.01 } \\                                                                                                                                         
 \packnet             &  0.22 \fs {   0.01 } &  \textbf{0.08} \fs {   0.01 } &  0.25  \fs {  0.00 } &  0.55 \fs {   0.01 } &  \textcolor{purple}{\textbf{0.07}}  \fs {   0.02 } &  \textcolor{purple}{\textbf{0.63}}  \fs {  0.00 } \\                                                                                                                                             
 \mtl                 &                      &                       &  0.48  \fs {  0.01 } &                      &                       &  0.83  \fs {  0.00 } \\
    \midrule 
    & \multicolumn{3}{c}{\liberoobject{}} & \multicolumn{3}{c}{\liberogoal{}} \\
    \cmidrule(lr){2-4} \cmidrule(l){5-7}
 \seql                &  \textbf{0.78} \fs {   0.04 } &  0.76  \fs {   0.04 } &  0.26  \fs {  0.02 }  &  \textcolor{purple}{\textbf{0.77}} \fs {   0.01 } &  0.82  \fs {   0.01 } &  0.22  \fs {  0.00 } \\
 \er                  &  0.67 \fs {   0.07 } &  0.43  \fs {   0.04 } &  0.44  \fs {  0.06 }  &  0.64 \fs {   0.01 } &  0.34  \fs {   0.02 } &  0.49  \fs {  0.02 } \\
 \ewc                 &  0.56 \fs {   0.03 } &  0.69  \fs {   0.02 } &  0.16  \fs {  0.02 }  &  0.32 \fs {   0.02 } &  0.48  \fs {   0.03 } &  0.06  \fs {  0.00 } \\
 \packnet             &  0.60 \fs {   0.07 } &  \textcolor{purple}{\textbf{0.17}}  \fs {   0.05 } &  \textbf{0.60}  \fs {  0.05 }  &  0.63 \fs {   0.02 } &  \textcolor{purple}{\textbf{0.06}}  \fs {   0.01 } &  \textcolor{purple}{\textbf{0.75}}  \fs {  0.01 } \\
 \mtl                 &                      &                       &  0.54  \fs {  0.02 }  &                      &                       &  0.80  \fs {  0.01 } \\
\bottomrule
    \end{tabular}
    }
    \vspace{5pt}
    \caption{Performance of three lifelong algorithms and the \seql{} and \mtl{} baselines on the four task suites, where the policy is fixed to be \bct{}. Results are averaged over three seeds and we report the mean and standard error. The best performance is \textbf{bolded}, and colored in \textcolor{purple}{\textbf{purple}} if the improvement is statistically significant over other algorithms, when a two-tailed, Student’s
t-test under equal sample sizes and unequal variance is applied with a $p$-value of 0.05.}
    \label{tab:algorithm-short}
\end{table}

\textcolor{purple}{\textit{Findings:}}
 We observed a series of interesting findings that could potentially benefit future research on algorithm design for \lldm{}: \textbf{1)} \seql{} shows the best FWT over all task suites. This is surprising since it indicates all lifelong learning algorithms we consider actually hurt forward transfer; \textbf{2)} \packnet{} outperforms other lifelong learning algorithms on \liberox{} but is outperformed by \er{} significantly on \liberolong{}, mainly because of low forward transfer. This confirms that the dynamic architecture approach is good at preventing forgetting. But since \packnet{} splits the network into different sub-networks, the essential capacity of the network for learning any individual task is smaller. Therefore, we conjecture that \packnet{} is not rich enough to learn on \liberolong{}; \textbf{3)} \ewc{} works worse than \seql{}, showing that the regularization on the loss term can actually impede the agent's performance on \lldm{} problems (See Appendix~\ref{appendix:loss-success-rates}); and \textbf{4)} \er{}, the rehearsal method, is robust across all task suites.

\myparagraph{Study on Language Embeddings as the Task Identifier (\qfour{})}
To investigate to what extent language embedding play a role in \lldm{}, we compare the performance of the same lifelong learner using four different pretrained language embeddings. Namely, we choose BERT~\citep{devlin2018bert}, CLIP~\citep{radford2021learning}, GPT-2~\citep{radford2019language} and the Task-ID embedding. Task-ID embeddings are produced by feeding a string such as “Task 5” into a pretrained BERT model.
  
\begin{table}[h!]
    \centering
    \begin{tabular}{ l c c c c}
    \toprule
    Embedding Type & Dimension & FWT($\uparrow$) & NBT($\downarrow$) & AUC($\uparrow$) \\
    \midrule 
    BERT    &  768 & 0.48 \fs {   0.02 } &  \textbf{0.32}  \fs {   0.04 } &  0.32  \fs {  0.01 } \\ 
    CLIP    &  512 &  \textbf{0.52} \fs {   0.00 } &  0.34  \fs {   0.01 } &  \textbf{0.35}  \fs {  0.01 } \\ 
    GPT-2    &  768 & 0.46 \fs {   0.01 } &  0.34  \fs {   0.02 } &  0.30  \fs {  0.01 } \\
    Task-ID &  768 & 0.50 \fs {   0.01 } &  0.37  \fs {   0.01 } &  0.33  \fs {  0.01 } \\
\bottomrule
    \end{tabular}
    \vspace{5pt}
    \caption{Performance of a lifelong learner using four different language embeddings on \liberolong{}, where we fix the policy architecture to \bct{} and the lifelong learning algorithm to \er{}. The Task-ID embeddings are retrieved by feeding ``Task + ID" into a pretrained BERT model. Results are averaged over three seeds and we report the mean and standard error. The best performance is \textbf{bolded}. No statistically significant difference is observed among the different language embeddings.}
    \label{tab:exp-language}
\end{table}

\textcolor{purple}{\textit{Findings:}} From Table~\ref{tab:exp-language}, 
we observe \emph{no} statistically significant difference among various language embeddings, including the Task-ID embedding. This, we believe, is due to sentence embeddings functioning as bag-of-words that differentiates different tasks.
This insight calls for better language encoding to harness the semantic information in task descriptions. Despite the similar performance, we opt for BERT embeddings as our default task embedding.

\begin{figure}[h!]
    \centering
    \includegraphics[width=0.7\textwidth]{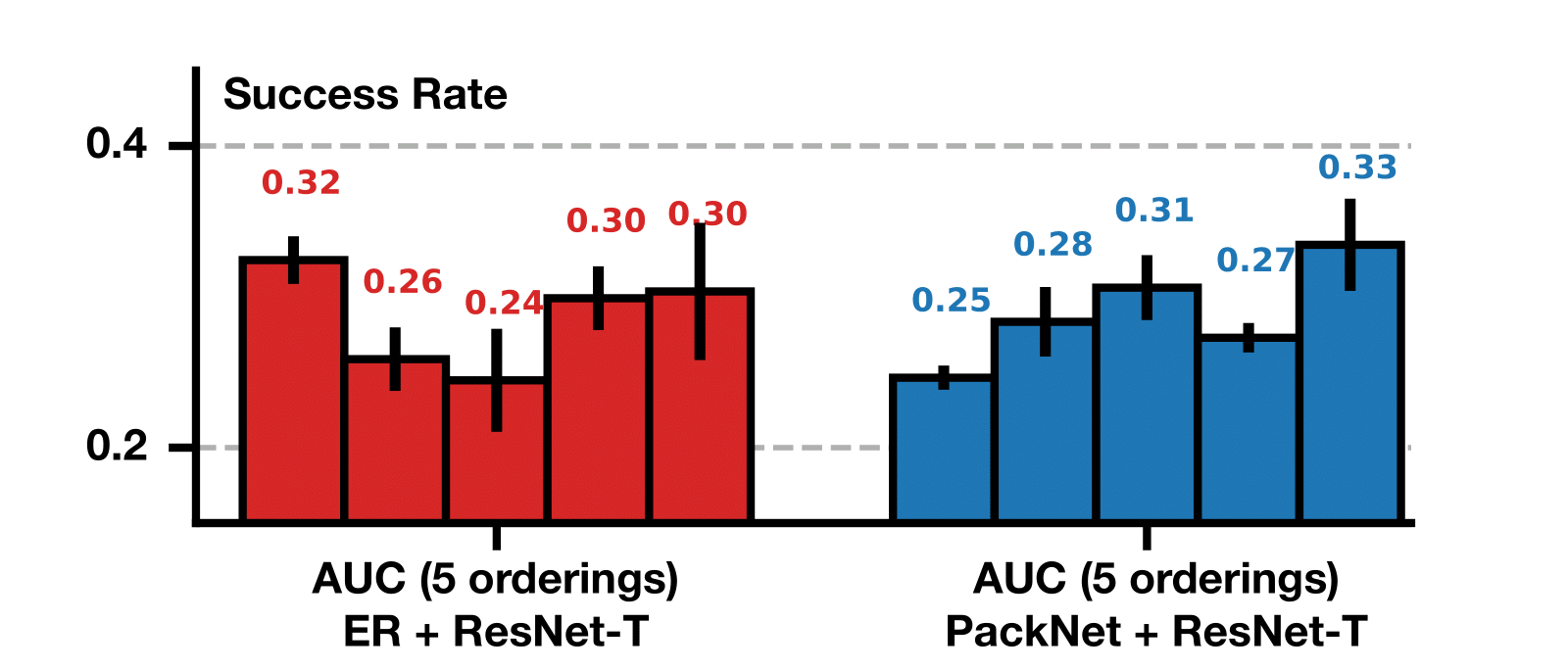}
    \caption{Performance of \er{} and \packnet{} using \bct{} on five different task orderings. An error bar shows the performance standard deviation for a fixed ordering.}
    \label{fig:exp-taskorder} 
\end{figure}

\myparagraph{Study on task ordering (\qfive{})}
Figure~\ref{fig:exp-taskorder} shows the result of the study on \qfour{}. For all experiments in this study, we used \bct{} as the neural architecture and evaluated both \er{} and \packnet{}. As the figure illustrates, the performance of both algorithms varies across different task orderings. This finding highlights an important direction for future research: developing algorithms or architectures that are robust to varying task orderings.

\textcolor{purple}{\textit{Findings:}} From Figure~\ref{fig:exp-taskorder}, we observe that indeed different task ordering could result in very different performances for the same algorithm. Specifically, such difference is statistically significant for \packnet{}.

\begin{figure*}[h!]
    \centering
    \includegraphics[width=\textwidth]{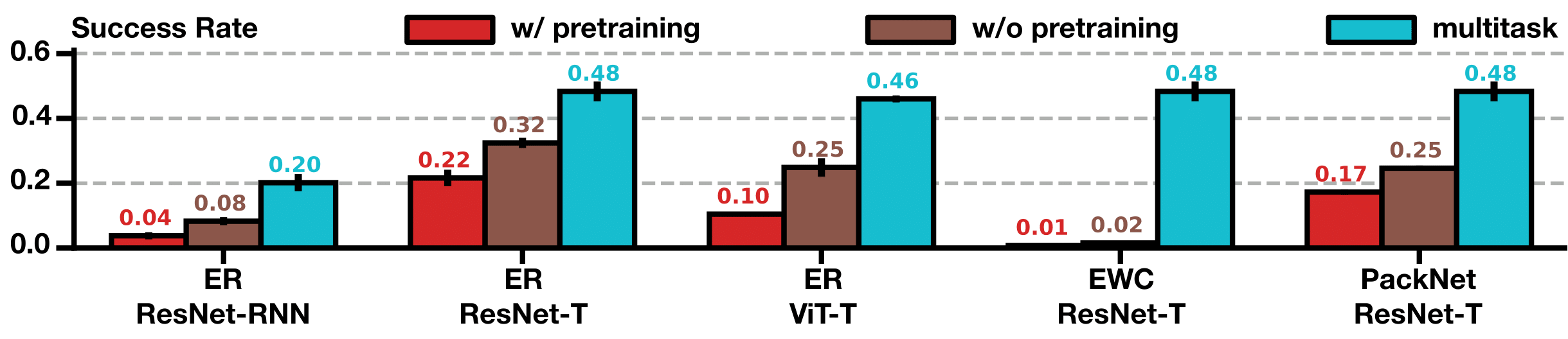}
    \caption{Performance of different combinations of algorithms and architectures without pretraining or with pretraining. The multi-task learning performance is also included for reference.}
    \label{fig:pretraining}
\end{figure*}

\myparagraph{Study on How Pretraining Affects Downstream \lldm{} (\qsix{})} Fig~\ref{fig:pretraining} reports the results on \liberolong{} of five combinations of algorithms and policy architectures, when the underlying model is pretrained on the 90 short-horizion tasks in \liberohundred{} or learned from scratch. For pretraining, we apply behavioral cloning on the 90 tasks using the three policy architectures for 50 epochs. We save a checkpoint every 5 epochs of training and then pick the checkpoint for each architecture that has the best performance as the pretrained model for downstream \lldm{}.

\textcolor{purple}{\textit{Findings:}} We observe that the basic supervised pretraining can \emph{hurt} the model's downstream lifelong learning performance. This, together with the results seen in Table~\ref{tab:algorithm-short} (e.g., naive sequential fine-tuning has better forward transfer than when lifelong learning algorithms are applied), indicates that better pretraining techniques are needed.

\myparagraph{Attention Visualization:} To better understand what type of knowledge the agent forgets during the lifelong learning process, we visualize the agent's attention map on each observed image input. The visualized saliency maps and the discussion can be found in Appendix~\ref{appendix:attention}.
\mysection{Related Work}
\label{sec:related}
This section provides an overview of existing benchmarks for lifelong learning and robot learning. We refer the reader to Appendix \ref{appendix:llalgo} for a detailed review of lifelong learning algorithms.

\myparagraph{Lifelong Learning Benchmarks} Pioneering work has adapted standard vision or language datasets for studying LL. This line of work includes image classification datasets like MNIST~\citep{deng2012mnist}, CIFAR~\citep{krizhevsky2009learning}, and ImageNet~\citep{deng2009imagenet}; segmentation datasets like Core50~\citep{lomonaco2017core50}; and natural language understanding datasets like GLUE~\citep{wang2018glue} and SuperGLUE~\citep{sarlin2020superglue}. Besides supervised learning datasets, video game benchmarks (e.g., Atari~\citep{mnih2013playing}, XLand~\citep{team2021open}, and VisDoom~\citep{kempka2016vizdoom}) in reinforcement learning (RL) have also been used for studying LL. However, LL in standard supervised learning does not involve procedural knowledge transfer, while RL problems in games do not represent human activities. ContinualWorld~\citep{Woczyk2021ContinualWA} modifies the 50 manipulation tasks in MetaWorld for LL. CORA~\citep{powers2021cora} builds four lifelong RL benchmarks based on Atari, Procgen~\citep{cobbe2020leveraging}, MiniHack~\citep{samvelyan2021minihack}, and ALFRED~\citep{shridhar2020alfred}. 
F-SIOL-310~\citep{ayub2021f} and OpenLORIS~\citep{she2020openloris} are challenging real-world lifelong object learning datasets that are captured from robotic vision systems. Prior works have also analyzed different components in a LL agent~\citep{mirzadeh2022architecture,Woczyk2022DisentanglingTI,ermis2022memory}, but they do not focus on robot manipulation problems.

\myparagraph{Robot Learning Benchmarks} A variety of robot learning benchmarks have been proposed to address challenges in meta learning (MetaWorld~\citep{yu2020meta}), causality learning (CausalWorld~\citep{ahmed2020causalworld}), multi-task learning~\citep{james2020rlbench, li2023behavior}, policy generalization to unseen objects~\citep{mu2021maniskill, gu2023maniskill2}, and compositional learning~\citep{mendez2022composuite}. Compared to existing benchmarks in lifelong learning and robot learning, the task suites in \lb{} are curated to address the research topics of \lldm{}. The benchmark includes a large number of tasks based on everyday human activities that feature rich interactive behaviors with a diverse range of objects. Additionally, the tasks in \lb{} are procedurally generated, making the benchmark scalable and adaptable. Moreover, the provided high-quality human demonstration dataset in \lb{} supports and encourages learning efficiency.
\mysection{Conclusion and Limitations}
This paper introduces \lb{}, a new benchmark in the robot manipulation domain for supporting research in \lldm{}. \lb{} includes a procedural generation pipeline that can create an infinite number of manipulation tasks in the simulator. We use this pipeline to create 130 standardized tasks and conduct a comprehensive set of experiments on policy and algorithm designs. The empirical results suggest several future research directions: 1) how to design a better neural architecture to better process spatial information or temporal information; 2) how to design a better algorithm to improve forward transfer ability; and 3) how to use pretraining to help improve lifelong learning performance.
In the short term, we do not envision any negative societal impacts triggered by \lb{}. But as the lifelong learner mainly learns from humans, studying how to preserve user privacy within \lldm{}~\citep{Liu2022ContinualLA} is crucial in the long run.

\bibliography{neurips_2023}

\begin{thebibliography}{10}

\bibitem{ahmed2020causalworld}
Ossama Ahmed, Frederik Tr{\"a}uble, Anirudh Goyal, Alexander Neitz, Yoshua
  Bengio, Bernhard Sch{\"o}lkopf, Manuel W{\"u}thrich, and Stefan Bauer.
\newblock Causalworld: A robotic manipulation benchmark for causal structure
  and transfer learning.
\newblock {\em arXiv preprint arXiv:2010.04296}, 2020.

\bibitem{ayub2022few}
Ali Ayub and Carter Fendley.
\newblock Few-shot continual active learning by a robot.
\newblock {\em arXiv preprint arXiv:2210.04137}, 2022.

\bibitem{ayub2021f}
Ali Ayub and Alan~R Wagner.
\newblock F-siol-310: A robotic dataset and benchmark for few-shot incremental
  object learning.
\newblock In {\em 2021 IEEE International Conference on Robotics and Automation
  (ICRA)}, pages 13496--13502. IEEE, 2021.

\bibitem{bain1995framework}
Michael Bain and Claude Sammut.
\newblock A framework for behavioural cloning.
\newblock In {\em Machine Intelligence 15}, pages 103--129, 1995.

\bibitem{ben2022lifelong}
Eseoghene Ben-Iwhiwhu, Saptarshi Nath, Praveen~K Pilly, Soheil Kolouri, and
  Andrea Soltoggio.
\newblock Lifelong reinforcement learning with modulating masks.
\newblock {\em arXiv preprint arXiv:2212.11110}, 2022.

\bibitem{bengio2009curriculum}
Yoshua Bengio, J{\'e}r{\^o}me Louradour, Ronan Collobert, and Jason Weston.
\newblock Curriculum learning.
\newblock In {\em Proceedings of the 26th annual international conference on
  machine learning}, pages 41--48, 2009.

\bibitem{biesialska2020continual}
Magdalena Biesialska, Katarzyna Biesialska, and Marta~R Costa-Jussa.
\newblock Continual lifelong learning in natural language processing: A survey.
\newblock {\em arXiv preprint arXiv:2012.09823}, 2020.

\bibitem{bishop1994mixture}
Christopher~M Bishop.
\newblock Mixture density networks.
\newblock 1994.

\bibitem{buzzega2020dark}
Pietro Buzzega, Matteo Boschini, Angelo Porrello, Davide Abati, and Simone
  Calderara.
\newblock Dark experience for general continual learning: a strong, simple
  baseline.
\newblock {\em Advances in neural information processing systems},
  33:15920--15930, 2020.

\bibitem{caruana1997multitask}
Rich Caruana.
\newblock Multitask learning.
\newblock {\em Machine learning}, 28(1):41--75, 1997.

\bibitem{chaudhry2018riemannian}
Arslan Chaudhry, Puneet~K Dokania, Thalaiyasingam Ajanthan, and Philip~HS Torr.
\newblock Riemannian walk for incremental learning: Understanding forgetting
  and intransigence.
\newblock In {\em Proceedings of the European Conference on Computer Vision
  (ECCV)}, pages 532--547, 2018.

\bibitem{chaudhry2018efficient}
Arslan Chaudhry, Marc'Aurelio Ranzato, Marcus Rohrbach, and Mohamed Elhoseiny.
\newblock Efficient lifelong learning with a-gem.
\newblock {\em arXiv preprint arXiv:1812.00420}, 2018.

\bibitem{chaudhry2019tiny}
Arslan Chaudhry, Marcus Rohrbach, Mohamed Elhoseiny, Thalaiyasingam Ajanthan,
  Puneet~K Dokania, Philip~HS Torr, and Marc'Aurelio Ranzato.
\newblock On tiny episodic memories in continual learning.
\newblock {\em arXiv preprint arXiv:1902.10486}, 2019.

\bibitem{cheung2019superposition}
Brian Cheung, Alexander Terekhov, Yubei Chen, Pulkit Agrawal, and Bruno
  Olshausen.
\newblock Superposition of many models into one.
\newblock {\em Advances in neural information processing systems}, 32, 2019.

\bibitem{cobbe2020leveraging}
Karl Cobbe, Chris Hesse, Jacob Hilton, and John Schulman.
\newblock Leveraging procedural generation to benchmark reinforcement learning.
\newblock In {\em International conference on machine learning}, pages
  2048--2056. PMLR, 2020.

\bibitem{de2021continual}
Matthias De~Lange, Rahaf Aljundi, Marc Masana, Sarah Parisot, Xu~Jia,
  Ale{\v{s}} Leonardis, Gregory Slabaugh, and Tinne Tuytelaars.
\newblock A continual learning survey: Defying forgetting in classification
  tasks.
\newblock {\em IEEE transactions on pattern analysis and machine intelligence},
  44(7):3366--3385, 2021.

\bibitem{deng2009imagenet}
Jia Deng, Wei Dong, Richard Socher, Li-Jia Li, Kai Li, and Li~Fei-Fei.
\newblock Imagenet: A large-scale hierarchical image database.
\newblock In {\em 2009 IEEE conference on computer vision and pattern
  recognition}, pages 248--255. Ieee, 2009.

\bibitem{deng2012mnist}
Li~Deng.
\newblock The mnist database of handwritten digit images for machine learning
  research.
\newblock {\em IEEE Signal Processing Magazine}, 29(6):141--142, 2012.

\bibitem{devlin2018bert}
Jacob Devlin, Ming-Wei Chang, Kenton Lee, and Kristina Toutanova.
\newblock Bert: Pre-training of deep bidirectional transformers for language
  understanding.
\newblock {\em arXiv preprint arXiv:1810.04805}, 2018.

\bibitem{diaz2018don}
Natalia D{\'\i}az-Rodr{\'\i}guez, Vincenzo Lomonaco, David Filliat, and Davide
  Maltoni.
\newblock Don't forget, there is more than forgetting: new metrics for
  continual learning.
\newblock {\em arXiv preprint arXiv:1810.13166}, 2018.

\bibitem{ermis2022memory}
Beyza Ermis, Giovanni Zappella, Martin Wistuba, and C{\'e}dric Archambeau.
\newblock Memory efficient continual learning with transformers.
\newblock 2022.

\bibitem{grauman2022ego4d}
Kristen Grauman, Andrew Westbury, Eugene Byrne, Zachary Chavis, Antonino
  Furnari, Rohit Girdhar, Jackson Hamburger, Hao Jiang, Miao Liu, Xingyu Liu,
  et~al.
\newblock Ego4d: Around the world in 3,000 hours of egocentric video.
\newblock In {\em Proceedings of the IEEE/CVF Conference on Computer Vision and
  Pattern Recognition}, pages 18995--19012, 2022.

\bibitem{Greydanus2017VisualizingAU}
Sam Greydanus, Anurag Koul, Jonathan Dodge, and Alan Fern.
\newblock Visualizing and understanding atari agents.
\newblock {\em ArXiv}, abs/1711.00138, 2017.

\bibitem{gu2023maniskill2}
Jiayuan Gu, Fanbo Xiang, Xuanlin Li, Zhan Ling, Xiqiang Liu, Tongzhou Mu, Yihe
  Tang, Stone Tao, Xinyue Wei, Yunchao Yao, et~al.
\newblock Maniskill2: A unified benchmark for generalizable manipulation
  skills.
\newblock {\em arXiv preprint arXiv:2302.04659}, 2023.

\bibitem{hausknecht2015deep}
Matthew Hausknecht and Peter Stone.
\newblock Deep recurrent q-learning for partially observable mdps.
\newblock In {\em 2015 aaai fall symposium series}, 2015.

\bibitem{hung2019compacting}
Ching-Yi Hung, Cheng-Hao Tu, Cheng-En Wu, Chien-Hung Chen, Yi-Ming Chan, and
  Chu-Song Chen.
\newblock Compacting, picking and growing for unforgetting continual learning.
\newblock {\em Advances in Neural Information Processing Systems}, 32, 2019.

\bibitem{james2020rlbench}
Stephen James, Zicong Ma, David~Rovick Arrojo, and Andrew~J Davison.
\newblock Rlbench: The robot learning benchmark \& learning environment.
\newblock {\em IEEE Robotics and Automation Letters}, 5(2):3019--3026, 2020.

\bibitem{kaelbling2020foundation}
Leslie~Pack Kaelbling.
\newblock The foundation of efficient robot learning.
\newblock {\em Science}, 369(6506):915--916, 2020.

\bibitem{kang2022class}
Minsoo Kang, Jaeyoo Park, and Bohyung Han.
\newblock Class-incremental learning by knowledge distillation with adaptive
  feature consolidation.
\newblock In {\em Proceedings of the IEEE/CVF conference on computer vision and
  pattern recognition}, pages 16071--16080, 2022.

\bibitem{kempka2016vizdoom}
Micha{\l} Kempka, Marek Wydmuch, Grzegorz Runc, Jakub Toczek, and Wojciech
  Ja{\'s}kowski.
\newblock Vizdoom: A doom-based ai research platform for visual reinforcement
  learning.
\newblock In {\em 2016 IEEE conference on computational intelligence and games
  (CIG)}, pages 1--8. IEEE, 2016.

\bibitem{kim2021vilt}
Wonjae Kim, Bokyung Son, and Ildoo Kim.
\newblock Vilt: Vision-and-language transformer without convolution or region
  supervision.
\newblock In {\em International Conference on Machine Learning}, pages
  5583--5594. PMLR, 2021.

\bibitem{kingma2014adam}
Diederik~P Kingma and Jimmy Ba.
\newblock Adam: A method for stochastic optimization.
\newblock {\em arXiv preprint arXiv:1412.6980}, 2014.

\bibitem{kirkpatrick2017overcoming}
James Kirkpatrick, Razvan Pascanu, Neil Rabinowitz, Joel Veness, Guillaume
  Desjardins, Andrei~A Rusu, Kieran Milan, John Quan, Tiago Ramalho, Agnieszka
  Grabska-Barwinska, et~al.
\newblock Overcoming catastrophic forgetting in neural networks.
\newblock {\em Proceedings of the national academy of sciences},
  114(13):3521--3526, 2017.

\bibitem{krizhevsky2009learning}
Alex Krizhevsky, Geoffrey Hinton, et~al.
\newblock Learning multiple layers of features from tiny images.
\newblock 2009.

\bibitem{li2023behavior}
Chengshu Li, Ruohan Zhang, Josiah Wong, Cem Gokmen, Sanjana Srivastava, Roberto
  Mart{\'\i}n-Mart{\'\i}n, Chen Wang, Gabrael Levine, Michael Lingelbach,
  Jiankai Sun, et~al.
\newblock Behavior-1k: A benchmark for embodied ai with 1,000 everyday
  activities and realistic simulation.
\newblock In {\em Conference on Robot Learning}, pages 80--93. PMLR, 2023.

\bibitem{Liu2022ContinualLA}
B.~Liu, Qian Liu, and Peter Stone.
\newblock Continual learning and private unlearning.
\newblock In {\em CoLLAs}, 2022.

\bibitem{liu2022continual}
Hao Liu and Huaping Liu.
\newblock Continual learning with recursive gradient optimization.
\newblock {\em arXiv preprint arXiv:2201.12522}, 2022.

\bibitem{lomonaco2017core50}
Vincenzo Lomonaco and Davide Maltoni.
\newblock Core50: a new dataset and benchmark for continuous object
  recognition.
\newblock In {\em Conference on Robot Learning}, pages 17--26. PMLR, 2017.

\bibitem{lopez2017gradient}
David Lopez-Paz and Marc'Aurelio Ranzato.
\newblock Gradient episodic memory for continual learning.
\newblock {\em Advances in neural information processing systems}, 30, 2017.

\bibitem{mai2022online}
Zheda Mai, Ruiwen Li, Jihwan Jeong, David Quispe, Hyunwoo Kim, and Scott
  Sanner.
\newblock Online continual learning in image classification: An empirical
  survey.
\newblock {\em Neurocomputing}, 469:28--51, 2022.

\bibitem{mallya2018packnet}
Arun Mallya and Svetlana Lazebnik.
\newblock Packnet: Adding multiple tasks to a single network by iterative
  pruning.
\newblock In {\em Proceedings of the IEEE conference on Computer Vision and
  Pattern Recognition}, pages 7765--7773, 2018.

\bibitem{mandlekar2021matters}
Ajay Mandlekar, Danfei Xu, Josiah Wong, Soroush Nasiriany, Chen Wang, Rohun
  Kulkarni, Li~Fei-Fei, Silvio Savarese, Yuke Zhu, and Roberto
  Mart{\'\i}n-Mart{\'\i}n.
\newblock What matters in learning from offline human demonstrations for robot
  manipulation.
\newblock {\em arXiv preprint arXiv:2108.03298}, 2021.

\bibitem{mcdermott1998pddl}
Drew McDermott, Malik Ghallab, Adele Howe, Craig Knoblock, Ashwin Ram, Manuela
  Veloso, Daniel Weld, and David Wilkins.
\newblock Pddl-the planning domain definition language.
\newblock 1998.

\bibitem{mendez2022composuite}
Jorge~A Mendez, Marcel Hussing, Meghna Gummadi, and Eric Eaton.
\newblock Composuite: A compositional reinforcement learning benchmark.
\newblock {\em arXiv preprint arXiv:2207.04136}, 2022.

\bibitem{mirzadeh2022architecture}
Seyed~Iman Mirzadeh, Arslan Chaudhry, Dong Yin, Timothy Nguyen, Razvan Pascanu,
  Dilan Gorur, and Mehrdad Farajtabar.
\newblock Architecture matters in continual learning.
\newblock {\em arXiv preprint arXiv:2202.00275}, 2022.

\bibitem{mnih2013playing}
Volodymyr Mnih, Koray Kavukcuoglu, David Silver, Alex Graves, Ioannis
  Antonoglou, Daan Wierstra, and Martin Riedmiller.
\newblock Playing atari with deep reinforcement learning.
\newblock {\em arXiv preprint arXiv:1312.5602}, 2013.

\bibitem{mu2021maniskill}
Tongzhou Mu, Zhan Ling, Fanbo Xiang, Derek Yang, Xuanlin Li, Stone Tao, Zhiao
  Huang, Zhiwei Jia, and Hao Su.
\newblock Maniskill: Generalizable manipulation skill benchmark with
  large-scale demonstrations.
\newblock {\em arXiv preprint arXiv:2107.14483}, 2021.

\bibitem{narvekar2020curriculum}
Sanmit Narvekar, Bei Peng, Matteo Leonetti, Jivko Sinapov, Matthew~E Taylor,
  and Peter Stone.
\newblock Curriculum learning for reinforcement learning domains: A framework
  and survey.
\newblock {\em arXiv preprint arXiv:2003.04960}, 2020.

\bibitem{parisi2019continual}
German~I Parisi, Ronald Kemker, Jose~L Part, Christopher Kanan, and Stefan
  Wermter.
\newblock Continual lifelong learning with neural networks: A review.
\newblock {\em Neural Networks}, 113:54--71, 2019.

\bibitem{perez2018film}
Ethan Perez, Florian Strub, Harm De~Vries, Vincent Dumoulin, and Aaron
  Courville.
\newblock Film: Visual reasoning with a general conditioning layer.
\newblock In {\em Proceedings of the AAAI Conference on Artificial
  Intelligence}, volume~32, 2018.

\bibitem{powers2021cora}
Sam Powers, Eliot Xing, Eric Kolve, Roozbeh Mottaghi, and Abhinav Gupta.
\newblock Cora: Benchmarks, baselines, and metrics as a platform for continual
  reinforcement learning agents.
\newblock {\em arXiv preprint arXiv:2110.10067}, 2021.

\bibitem{radford2021learning}
Alec Radford, Jong~Wook Kim, Chris Hallacy, Aditya Ramesh, Gabriel Goh,
  Sandhini Agarwal, Girish Sastry, Amanda Askell, Pamela Mishkin, Jack Clark,
  et~al.
\newblock Learning transferable visual models from natural language
  supervision.
\newblock In {\em International conference on machine learning}, pages
  8748--8763. PMLR, 2021.

\bibitem{radford2019language}
Alec Radford, Jeffrey Wu, Rewon Child, David Luan, Dario Amodei, Ilya
  Sutskever, et~al.
\newblock Language models are unsupervised multitask learners.
\newblock {\em OpenAI blog}, 1(8):9, 2019.

\bibitem{rios2020lifelong}
Amanda Rios and Laurent Itti.
\newblock Lifelong learning without a task oracle.
\newblock In {\em 2020 IEEE 32nd International Conference on Tools with
  Artificial Intelligence (ICTAI)}, pages 255--263. IEEE, 2020.

\bibitem{ross2011reduction}
St{\'e}phane Ross, Geoffrey Gordon, and Drew Bagnell.
\newblock A reduction of imitation learning and structured prediction to
  no-regret online learning.
\newblock In {\em Proceedings of the fourteenth international conference on
  artificial intelligence and statistics}, pages 627--635. JMLR Workshop and
  Conference Proceedings, 2011.

\bibitem{rusu2016progressive}
Andrei~A Rusu, Neil~C Rabinowitz, Guillaume Desjardins, Hubert Soyer, James
  Kirkpatrick, Koray Kavukcuoglu, Razvan Pascanu, and Raia Hadsell.
\newblock Progressive neural networks.
\newblock {\em arXiv preprint arXiv:1606.04671}, 2016.

\bibitem{saha2021space}
Gobinda Saha, Isha Garg, Aayush Ankit, and Kaushik Roy.
\newblock Space: Structured compression and sharing of representational space
  for continual learning.
\newblock {\em IEEE Access}, 9:150480--150494, 2021.

\bibitem{samvelyan2021minihack}
Mikayel Samvelyan, Robert Kirk, Vitaly Kurin, Jack Parker-Holder, Minqi Jiang,
  Eric Hambro, Fabio Petroni, Heinrich K{\"u}ttler, Edward Grefenstette, and
  Tim Rockt{\"a}schel.
\newblock Minihack the planet: A sandbox for open-ended reinforcement learning
  research.
\newblock {\em arXiv preprint arXiv:2109.13202}, 2021.

\bibitem{sarlin2020superglue}
Paul-Edouard Sarlin, Daniel DeTone, Tomasz Malisiewicz, and Andrew Rabinovich.
\newblock Superglue: Learning feature matching with graph neural networks.
\newblock In {\em Proceedings of the IEEE/CVF conference on computer vision and
  pattern recognition}, pages 4938--4947, 2020.

\bibitem{schwarz2018progress}
Jonathan Schwarz, Wojciech Czarnecki, Jelena Luketina, Agnieszka
  Grabska-Barwinska, Yee~Whye Teh, Razvan Pascanu, and Raia Hadsell.
\newblock Progress \& compress: A scalable framework for continual learning.
\newblock In {\em International Conference on Machine Learning}, pages
  4528--4537. PMLR, 2018.

\bibitem{she2020openloris}
Qi~She, Fan Feng, Xinyue Hao, Qihan Yang, Chuanlin Lan, Vincenzo Lomonaco,
  Xuesong Shi, Zhengwei Wang, Yao Guo, Yimin Zhang, et~al.
\newblock Openloris-object: A robotic vision dataset and benchmark for lifelong
  deep learning.
\newblock In {\em 2020 IEEE international conference on robotics and automation
  (ICRA)}, pages 4767--4773. IEEE, 2020.

\bibitem{shridhar2020alfred}
Mohit Shridhar, Jesse Thomason, Daniel Gordon, Yonatan Bisk, Winson Han,
  Roozbeh Mottaghi, Luke Zettlemoyer, and Dieter Fox.
\newblock Alfred: A benchmark for interpreting grounded instructions for
  everyday tasks.
\newblock In {\em Proceedings of the IEEE/CVF conference on computer vision and
  pattern recognition}, pages 10740--10749, 2020.

\bibitem{srivastava2022behavior}
Sanjana Srivastava, Chengshu Li, Michael Lingelbach, Roberto
  Mart{\'\i}n-Mart{\'\i}n, Fei Xia, Kent~Elliott Vainio, Zheng Lian, Cem
  Gokmen, Shyamal Buch, Karen Liu, et~al.
\newblock Behavior: Benchmark for everyday household activities in virtual,
  interactive, and ecological environments.
\newblock In {\em Conference on Robot Learning}, pages 477--490. PMLR, 2022.

\bibitem{team2021open}
Open Ended~Learning Team, Adam Stooke, Anuj Mahajan, Catarina Barros, Charlie
  Deck, Jakob Bauer, Jakub Sygnowski, Maja Trebacz, Max Jaderberg, Michael
  Mathieu, et~al.
\newblock Open-ended learning leads to generally capable agents.
\newblock {\em arXiv preprint arXiv:2107.12808}, 2021.

\bibitem{thrun1995lifelong}
Sebastian Thrun and Tom~M Mitchell.
\newblock Lifelong robot learning.
\newblock {\em Robotics and autonomous systems}, 15(1-2):25--46, 1995.

\bibitem{vaswani2017attention}
Ashish Vaswani, Noam Shazeer, Niki Parmar, Jakob Uszkoreit, Llion Jones,
  Aidan~N Gomez, {\L}ukasz Kaiser, and Illia Polosukhin.
\newblock Attention is all you need.
\newblock {\em Advances in neural information processing systems}, 30, 2017.

\bibitem{wang2018glue}
Alex Wang, Amanpreet Singh, Julian Michael, Felix Hill, Omer Levy, and Samuel~R
  Bowman.
\newblock Glue: A multi-task benchmark and analysis platform for natural
  language understanding.
\newblock {\em arXiv preprint arXiv:1804.07461}, 2018.

\bibitem{wang2023mimicplay}
Chen Wang, Linxi Fan, Jiankai Sun, Ruohan Zhang, Li~Fei-Fei, Danfei Xu, Yuke
  Zhu, and Anima Anandkumar.
\newblock Mimicplay: Long-horizon imitation learning by watching human play.
\newblock {\em arXiv preprint arXiv:2302.12422}, 2023.

\bibitem{Woczyk2021ContinualWA}
Maciej Wołczyk, Michal Zajkac, Razvan Pascanu, Lukasz Kuci'nski, and Piotr
  Milo's.
\newblock Continual world: A robotic benchmark for continual reinforcement
  learning.
\newblock In {\em Neural Information Processing Systems}, 2021.

\bibitem{Woczyk2022DisentanglingTI}
Maciej Wołczyk, Michal Zajkac, Razvan Pascanu, Lukasz Kuci'nski, and Piotr
  Milo's.
\newblock Disentangling transfer in continual reinforcement learning.
\newblock {\em ArXiv}, abs/2209.13900, 2022.

\bibitem{wu2020firefly}
Lemeng Wu, Bo~Liu, Peter Stone, and Qiang Liu.
\newblock Firefly neural architecture descent: a general approach for growing
  neural networks.
\newblock {\em Advances in Neural Information Processing Systems},
  33:22373--22383, 2020.

\bibitem{yoon2017lifelong}
Jaehong Yoon, Eunho Yang, Jeongtae Lee, and Sung~Ju Hwang.
\newblock Lifelong learning with dynamically expandable networks.
\newblock {\em arXiv preprint arXiv:1708.01547}, 2017.

\bibitem{yu2020meta}
Tianhe Yu, Deirdre Quillen, Zhanpeng He, Ryan Julian, Karol Hausman, Chelsea
  Finn, and Sergey Levine.
\newblock Meta-world: A benchmark and evaluation for multi-task and meta
  reinforcement learning.
\newblock In {\em Conference on robot learning}, pages 1094--1100. PMLR, 2020.

\bibitem{zhou2022forward}
Da-Wei Zhou, Fu-Yun Wang, Han-Jia Ye, Liang Ma, Shiliang Pu, and De-Chuan Zhan.
\newblock Forward compatible few-shot class-incremental learning.
\newblock In {\em Proceedings of the IEEE/CVF Conference on Computer Vision and
  Pattern Recognition}, pages 9046--9056, 2022.

\bibitem{zhu2022viola}
Yifeng Zhu, Abhishek Joshi, Peter Stone, and Yuke Zhu.
\newblock Viola: Imitation learning for vision-based manipulation with object
  proposal priors.
\newblock {\em arXiv preprint arXiv:2210.11339}, 2022.

\bibitem{zhu2020robosuite}
Yuke Zhu, Josiah Wong, Ajay Mandlekar, and Roberto Mart{\'\i}n-Mart{\'\i}n.
\newblock robosuite: A modular simulation framework and benchmark for robot
  learning.
\newblock {\em arXiv preprint arXiv:2009.12293}, 2020.

\end{thebibliography}
\bibliographystyle{plain}

\newpage
\section*{Checklist}
\begin{enumerate}
\item For all authors...
\begin{enumerate}
  \item Do the main claims made in the abstract and introduction accurately reflect the paper's contributions and scope?
    \answerYes{}
  \item Did you describe the limitations of your work?
    \answerYes{}
  \item Did you discuss any potential negative societal impacts of your work?
    \answerYes{}
  \item Have you read the ethics review guidelines and ensured that your paper conforms to them?
    \answerYes{}
\end{enumerate}

\item If you are including theoretical results...
\begin{enumerate}
  \item Did you state the full set of assumptions of all theoretical results?
    \answerNA{}
	\item Did you include complete proofs of all theoretical results?
    \answerNA{}
\end{enumerate}

\item If you ran experiments (e.g. for benchmarks)...
\begin{enumerate}
  \item Did you include the code, data, and instructions needed to reproduce the main experimental results (either in the supplemental material or as a URL)?
    \answerYes{}
  \item Did you specify all the training details (e.g., data splits, hyperparameters, how they were chosen)?
    \answerYes{}
	\item Did you report error bars (e.g., with respect to the random seed after running experiments multiple times)?
    \answerYes{}
	\item Did you include the total amount of compute and the type of resources used (e.g., type of GPUs, internal cluster, or cloud provider)?
    \answerYes{}
\end{enumerate}

\item If you are using existing assets (e.g., code, data, models) or curating/releasing new assets...
\begin{enumerate}
  \item If your work uses existing assets, did you cite the creators?
    \answerYes{}
  \item Did you mention the license of the assets?
    \answerNA{}
  \item Did you include any new assets either in the supplemental material or as a URL?
    \answerYes{}
  \item Did you discuss whether and how consent was obtained from people whose data you're using/curating?
    \answerNA{}
  \item Did you discuss whether the data you are using/curating contains personally identifiable information or offensive content?
    \answerNA{}
\end{enumerate}

\item If you used crowdsourcing or conducted research with human subjects...
\begin{enumerate}
  \item Did you include the full text of instructions given to participants and screenshots, if applicable?
    \answerNA{}
  \item Did you describe any potential participant risks, with links to Institutional Review Board (IRB) approvals, if applicable?
    \answerNA{}
  \item Did you include the estimated hourly wage paid to participants and the total amount spent on participant compensation?
    \answerNA{}
\end{enumerate}

\end{enumerate}
\newpage
\appendix

\section{Implemented Neural Architectures and Lifelong Learning Algorithms}

\begin{table}[h!]
    \centering
    \begin{tabular}{ l l}
    \toprule
    \multirow{3}{*}{Neural Policy Arch.} &  \bcrnn{}  \\
                                         &  \bct{}  \\
                                         &  \bcvilt{}  \\
    \midrule 
    \multirow{5}{*}{Lifelong Learning Algo.} &  \seql{}  \\
                                             &  \ewc{}~\citep{kirkpatrick2017overcoming}  \\
                                             &  \er{}~\citep{chaudhry2019tiny}  \\
                                             &  \packnet{}~\citep{mallya2018packnet}  \\
                                             &  \mtl{}  \\
                                    
\bottomrule
    \end{tabular}
    \caption{The implemented neural policy architectures and the lifelong learning algorithms in \lb{}.}
    \label{tab:arch-policy}
\end{table}
\subsection{Neural Architectures}
\label{appendix:neural}

\begin{figure*}[h!]
    \centering
    \includegraphics[width=\textwidth]{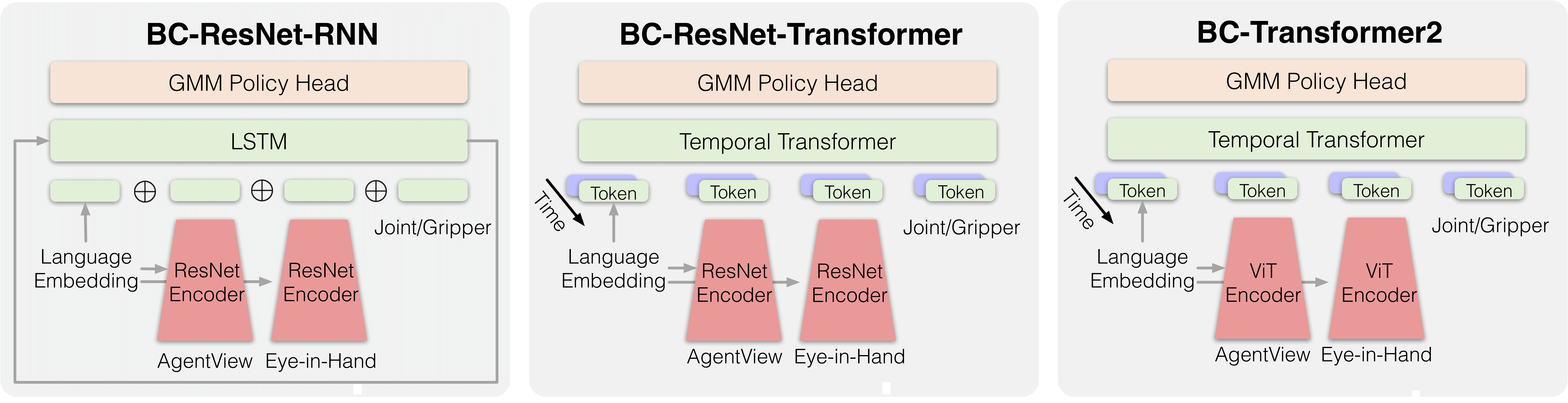}
    \caption{We provide visualizations of the architectures for \bcrnn{}, \bct{}, and \bcvilt{}, respectively. It is worth noting that each model architecture incorporates language embedding in distinct ways.}
    \label{fig:architectures}
\end{figure*}

In Section~\ref{sec:method-architecture}, we outlined the neural network architectures utilized in our experiments, namely \bcrnn{}, \bct{}, and \bcvilt{}. The specifics of each architecture are illustrated in Figure~\ref{fig:architectures}. Furthermore, Table~\ref{tab:rnn-param}, \ref{tab:transformer-param}, and \ref{tab:vilt-param} display the hyperparameters for the architectures used throughout all of our experiments.




\begin{figure}[h!]
    \centering
    \includegraphics[width=0.5\textwidth]{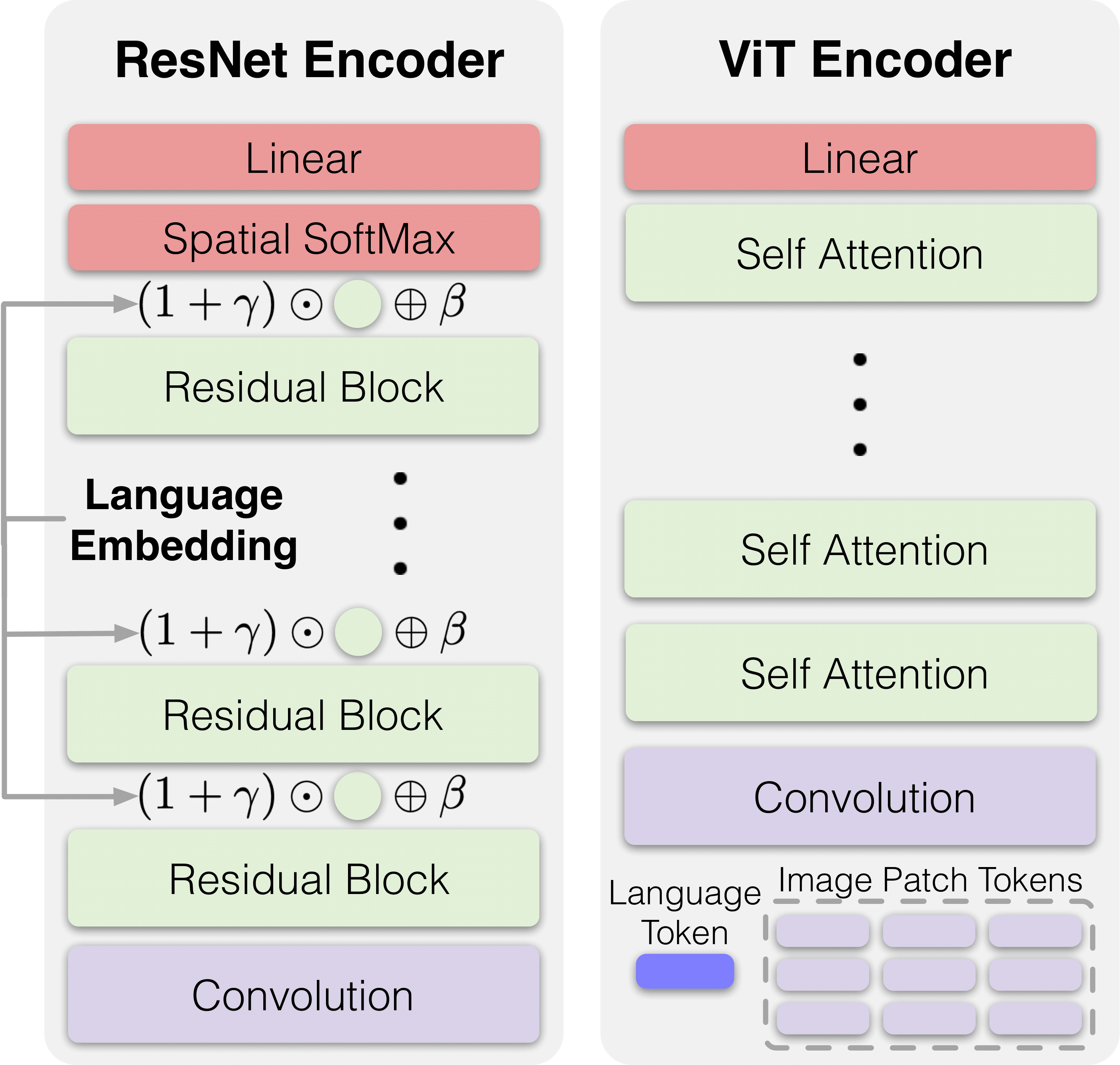}
    \caption{The image encoders: ResNet-based encoder and the vision transformer-based encoder.}
    \label{fig:encoders}
\end{figure}

\section{Computation}
For all experiments, we use a single Nvidia A100 GPU or a single Nvidia A40 GPU (CUDA 11.7) with 8~16 CPUs for training and evaluation.

\newpage

\begin{minipage}{\textwidth}
\begin{minipage}[h!]{0.48\textwidth}
\makeatletter\def\@captype{table}
\centering
\begin{tabular}{cc}
    \toprule
    Variable &  Value \\
    \midrule
 resnet\_image\_embed\_size & 64 \\
 text\_embed\_size & 32 \\
 rnn\_hidden\_size & 1024 \\
 rnn\_layer\_num & 2 \\
 rnn\_dropout & 0.0 \\
    \bottomrule
\end{tabular}
    \caption{Hyper parameters of \bcrnn{}.}
    \label{tab:rnn-param}
\end{minipage}
\begin{minipage}[t]{0.48\textwidth}
\makeatletter\def\@captype{table}
\centering
\begin{tabular}{cc}
    \toprule
    Variable &  Value \\
    \midrule
extra\_info\_hidden\_size & 128 \\
img\_embed\_size & 64 \\
transformer\_num\_layers & 4 \\
transformer\_num\_heads & 6 \\
transformer\_head\_output\_size & 64 \\
transformer\_mlp\_hidden\_size & 256 \\
transformer\_dropout & 0.1 \\
transformer\_max\_seq\_len & 10 \\
    \bottomrule
\end{tabular}
    \caption{Hyper parameters of \bct{}.}
    \label{tab:transformer-param}
\end{minipage}
\end{minipage}

\begin{table}[h!]
    \centering
    \begin{tabular}{cc}
    \toprule
    Variable &  Value \\
    \midrule
extra\_info\_hidden\_size & 128 \\
img\_embed\_size & 128 \\
spatial\_transformer\_num\_layers & 7 \\
spatial\_transformer\_num\_heads & 8 \\
spatial\_transformer\_head\_output\_size & 120 \\
spatial\_transformer\_mlp\_hidden\_size & 256 \\
spatial\_transformer\_dropout & 0.1 \\
spatial\_down\_sample\_embed\_size & 64 \\
temporal\_transformer\_input\_size & null \\
temporal\_transformer\_num\_layers & 4 \\
temporal\_transformer\_num\_heads & 6 \\
temporal\_transformer\_head\_output\_size & 64 \\
temporal\_transformer\_mlp\_hidden\_size & 256 \\
temporal\_transformer\_dropout & 0.1 \\
temporal\_transformer\_max\_seq\_len & 10 \\
    \bottomrule 
    \end{tabular}
\caption{Hyper parameters of \bcvilt{}.}
\label{tab:vilt-param}
\end{table}

\subsection{Lifelong Learning Algorithms}
\label{appendix:llalgo}
Lifelong learning (LL) is a field of study that aims to understand how an agent can continually acquire and retain knowledge over an infinite sequence of tasks without catastrophically forgetting previous knowledge. Recent literature proposes three main approaches to address the problem of catastrophic forgetting in deep learning: Dynamic Architecture approaches, Regularization-Based approaches, and Rehearsal approaches. Although some recent works explore the combination of different approaches~\citep{ayub2022few,kang2022class,rios2020lifelong} or new strategies~\citep{zhou2022forward,saha2021space,cheung2019superposition}, our benchmark aims to provide an in-depth analysis of these three basic lifelong learning directions to reveal their pros and cons on robot learning tasks.

The dynamic architecture approach gradually expands the learning model to incorporate new knowledge~\citep{rusu2016progressive,yoon2017lifelong,mallya2018packnet,hung2019compacting,wu2020firefly,ben2022lifelong}. Regularization-based methods, on the other hand, regularize the learner to a previous checkpoint when it learns a new task~\citep{kirkpatrick2017overcoming,chaudhry2018riemannian,schwarz2018progress,liu2022continual}. Rehearsal methods save exemplar data from prior tasks and replay them with new data to consolidate the agent's memory~\citep{chaudhry2019tiny,lopez2017gradient,chaudhry2018efficient,buzzega2020dark}. For a comprehensive review of LL methods, we refer readers to surveys \citep{de2021continual, parisi2019continual}.


The following paragraphs provide details on the three lifelong learning algorithms that we have implemented.

\paragraph{\er{}} Experience Replay (\er{})~\citep{chaudhry2019tiny} is a \textbf{rehearsal-based} approach that maintains a memory buffer of samples from previous tasks and leverages it to learn new tasks. After the completion of policy learning for a task, \er{} stores a portion of the data into a storage memory. When training a new task, \er{} samples data from the memory and combines it with the training data from the current task so that the training data approximately represents the empirical distribution of all-task data. In our implementation, we use a replay buffer to store a portion of the training data (up to 1000 trajectories) after training each task. For every training iteration during the training of a new task, we uniformly sample a fixed number of replay data from the memory (32 trajectories) along with each batch of training data from the new task.

\paragraph{\ewc{}} Elastic Weight Consolidation(\ewc{})~\citep{kirkpatrick2017overcoming} is a \textbf{regularization-based} approach that add a regularization term that constraints neural network update to the original single-task learning objective. Specifically, \ewc{} uses the Fisher information matrix that quantify the importance of every neural netwrk parameter. The loss function for task $k$ is:
$$
\mathcal{L}_{k}^{EWC}(\theta)=\mathcal{L}_K^{BC}(\theta)+\sum_i \frac{\lambda}{2} F_i\left(\theta_i-\theta_{{k-1}, i}^*\right)^2,
$$
where $\lambda$ is a penalty hyperparameter, and the coefficient $F_i$ is the diagonal of the Fisher information matrix: $F_k=\mathbb{E}_{s \sim \mathcal{D}_k} \mathbb{E}_{a \sim p_\theta(\cdot \mid s)}\left(\nabla_{\theta_k} \log p_{\theta_k}(a | s)\right)^2$. In this work, we use the online update version of \ewc{} that updates the Fisher information matrix using  exponential moving average along the lifelong learning process, and use the empirical estimation of above Fisher information matrix to stabilize the estimation. Formally, the actually used estimation of Fisher Information Matrix is $\tilde{F_k}=\gamma F_{k-1} + (1-\gamma) F_k$, where $F_k=\mathbb{E}_{(s,a) \sim \mathcal{D}_k} \left(\nabla_{\theta_k} \log p_{\theta_k}(a | s)\right)^2$ and $k$ is the task number. We set $\gamma=0.9$ and $\lambda=5 \cdot 10^4$.

\paragraph{\packnet{}}
\packnet{}~\citep{mallya2018packnet} is a \textbf{dynamic architecture-based} approach that aims to prevent changes to parameters that are important for previous tasks in lifelong learning. To achieve this, \packnet{} iteratively trains, prunes, fine-tunes, and freezes parts of the network. The method theoretically completely avoids catastrophic forgetting, but for each new task, the number of available parameters shrinks. The pruning process in \packnet{} involves two stages. First, the network is trained, and at the end of the training, a fixed proportion of the most important parameters (25\% in our implementation) are chosen, and the rest are pruned. Second, the selected part of the network is fine-tuned and then frozen. In our implementation, we follow the original paper \citep{mallya2018packnet} and do not train all biases and normalization layers. We perform the same number of fine-tuning epochs as for training (50 epochs in our implementation). Note that all evaluation metrics are calculated \textit{before} the fine-tuning stage.

\clearpage

\section{LIBERO Task Suite Designs}
\label{appendix:task}

\subsection{Task Suites}
\begin{figure}[h!]
    \centering
    \includegraphics[width=0.8\linewidth]{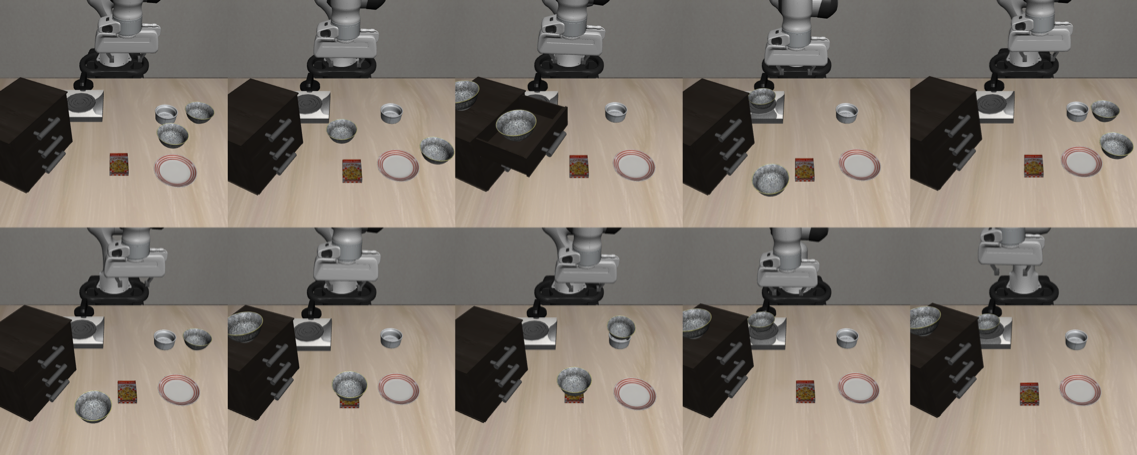}
    \caption{\liberospatial{}}
    \label{fig:libero-spatial}
\end{figure}

\begin{figure}[h!]
    \centering
    \includegraphics[width=0.8\linewidth]{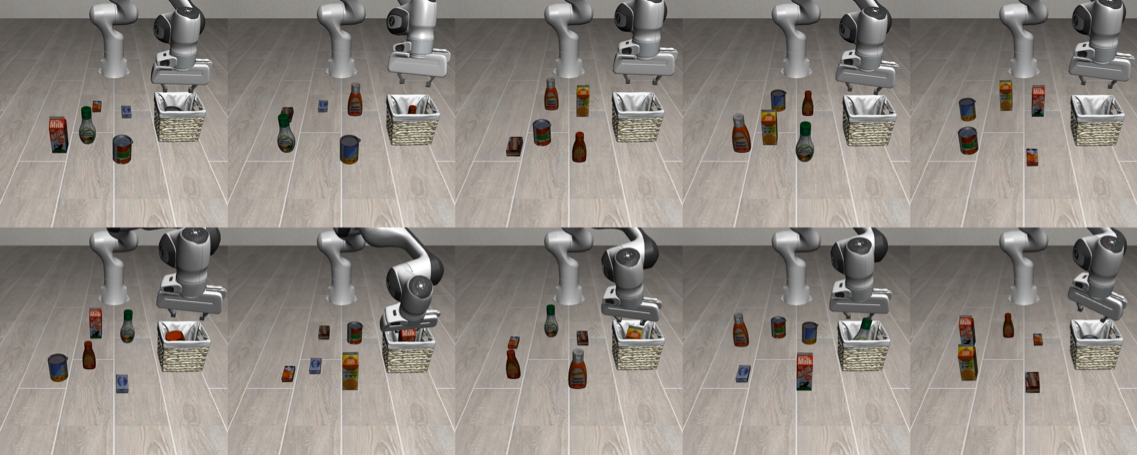}
    \caption{\liberoobject{}}
    \label{fig:libero-object}
\end{figure}

\begin{figure}[h!]
    \centering
    \includegraphics[width=0.8\linewidth]{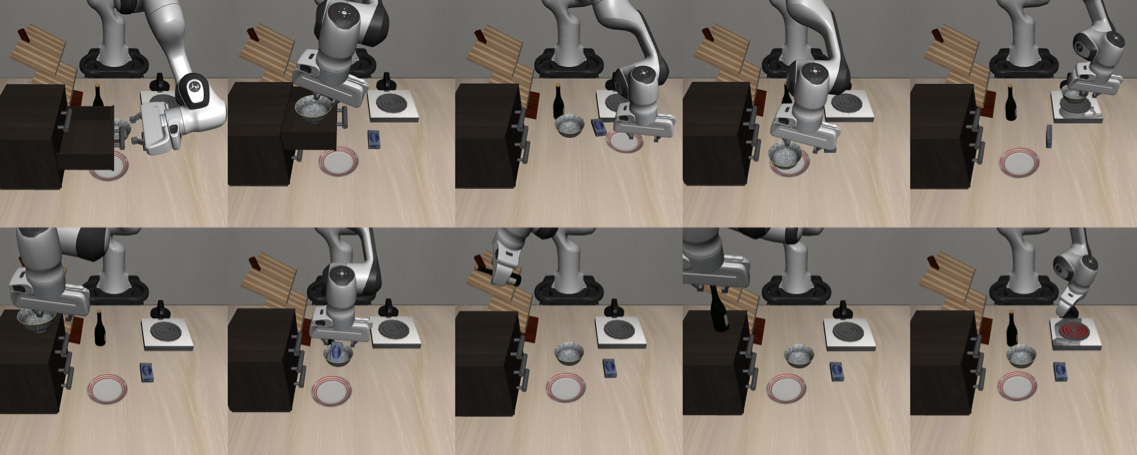}
    \caption{\liberogoal{}}
    \label{fig:libero-goal}
\end{figure}

\begin{figure}[h!]
    \centering
    \includegraphics[width=0.9\linewidth]{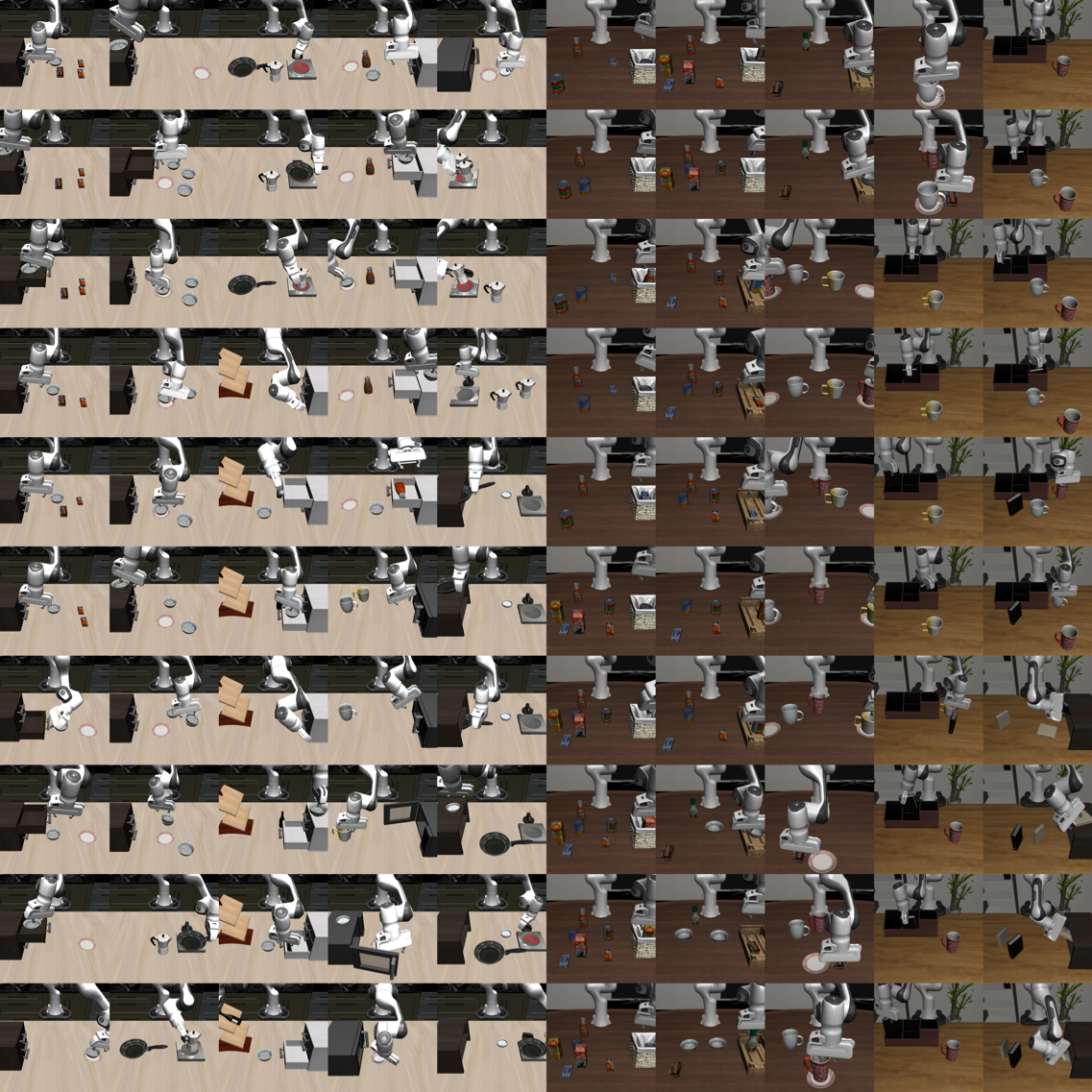}
    \caption{\liberohundred{}}
    \label{fig:libero-100}
\end{figure}

We visualize all the tasks from the four task suites in Figure~\ref{fig:libero-spatial}-~\ref{fig:libero-100}. Figure~\ref{fig:libero-spatial} visualizes the initial states since the task goals are always the same. All the figures visualize the goal states of tasks except for Figure~\ref{fig:libero-spatial}, which visualizes the initial states since the task goals are always the same.

~\\
~\\
~\\
~\\
~\\
~\\
~\\

\clearpage
\subsection{PDDL-based Scene Description File}
Here we visualize the whole content of an example scene description file based on PDDL. This file corresponds to the task shown in Figure~\ref{fig:libero-procedural-generation}.

\begin{center}
    \textbf{Example task:}\qquad \textcolor{purple}{\textit{Open the top drawer of the cabinet and put the bowl in it}}.
\end{center}

\begin{lstlisting}[caption={},label=lst:pddl-example]
(define (problem LIBERO_Kitchen_Tabletop_Manipulation)
  (:domain robosuite)
  (:language open the top drawer of the cabinet and put the bowl in it)
    (:regions
      (wooden_cabinet_init_region
          (:target kitchen_table)
          (:ranges (
              (-0.01 -0.31 0.01 -0.29)
            )
          )
          (:yaw_rotation (
              (3.141592653589793 3.141592653589793)
            )
          )
      )
      (akita_black_bowl_init_region
          (:target kitchen_table)
          (:ranges (
              (-0.025 -0.025 0.025 0.025)
            )
          )
          (:yaw_rotation (
              (0.0 0.0)
            )
          )
      )
      (plate_init_region
          (:target kitchen_table)
          (:ranges (
              (-0.025 0.225 0.025 0.275)
            )
          )
          (:yaw_rotation (
              (0.0 0.0)
            )
          )
      )
      (top_side
          (:target wooden_cabinet_1)
      )
      (top_region
          (:target wooden_cabinet_1)
      )
      (middle_region
          (:target wooden_cabinet_1)
      )
      (bottom_region
          (:target wooden_cabinet_1)
      )
    )

  (:fixtures
    kitchen_table - kitchen_table
    wooden_cabinet_1 - wooden_cabinet
  )

  (:objects
    akita_black_bowl_1 - akita_black_bowl
    plate_1 - plate
  )

  (:obj_of_interest
    wooden_cabinet_1
    akita_black_bowl_1
  )

  (:init
    (On akita_black_bowl_1 kitchen_table_akita_black_bowl_init_region)
    (On plate_1 kitchen_table_plate_init_region)
    (On wooden_cabinet_1 kitchen_table_wooden_cabinet_init_region)
  )

  (:goal
    (And (Open wooden_cabinet_1_top_region) 
         (In akita_black_bowl_1 wooden_cabinet_1_top_region)
    )
  )

)
\end{lstlisting}

\newpage

\section{Experimental Setup}
\label{appendix:exp-setting}

We consider five lifelong learning algorithms: \seql{} the sequential learning baseline where the agent learns each task in the sequence directly without any further consideration, \mtl{} the multitask learning baseline where the agent learns all tasks in the sequence simultaneously, the regularization-based method \ewc{}~\citep{kirkpatrick2017overcoming}, the replay-based method \er{}~\citep{chaudhry2019tiny}, and the dynamic architecture-based method \packnet{}~\citep{mallya2018packnet}. \seql{} and \mtl{} can be seen as approximations of the lower and upper bounds respectively for any lifelong learning algorithm. The other three methods represent the three primary categories of lifelong learning algorithms.
For the neural architectures, we consider three vision-language policy architectures: \bcrnn{}, \bct{}, \bcvilt{}, which differ in how spatial or temporal information is aggregated (See Appendix~\ref{appendix:neural} for more details).
For each task, the agent is trained over 50 epochs on the 50 demonstration trajectories. We evaluate the agent's average success rate over 20 test rollout trajectories of a maximum length of 600 every 5 epochs. We use Adam optimizer~\citep{kingma2014adam} with a batch size of $32$, and a cosine scheduled learning rate from $0.0001$ to $0.00001$ for each task.
Following the convention of~\robomimic{}~\citep{mandlekar2021matters}, we pick the model checkpoint that achieves the best success rate as the final policy for a given task.
After 50 epochs of training, the agent with the best checkpoint is then evaluated on all previously learned tasks, with 20 test rollout trajectories for each task. All policy networks are matched in Floating Point Operations Per Second (FLOPS): all policy architectures have $\sim$13.5G FLOPS. For each combination of algorithm, policy architecture, and task suite, we run the lifelong learning method 3 times with random seeds $\{100, 200, 300\}$ (180 experiments in total). See Table~\ref{tab:arch-policy} for the implemented algorithms and architectures.

\section{Additional Experiment Results}
\label{appendix:additional}

\subsection{Full Results}
\label{appendix:additional-result}

We provide the full results across three different lifelong learning algorithms (e.g., \ewc{}, \er{}, \packnet{}) and three different policy architectures (e.g., \bcrnn{}, \bct{}, \bcvilt{}) on the four task suites in Table~\ref{tab:algorithm}.

\begin{table}[h!]
    \centering
    \resizebox{\textwidth}{!}{
    \begin{tabular}{l l  c c c c c c}
    \toprule
 Algo. &   Policy Arch. & FWT($\uparrow$) & NBT($\downarrow$) & AUC($\uparrow$) & FWT($\uparrow$) & NBT($\downarrow$) & AUC($\uparrow$)\\
    \midrule 
    & & \multicolumn{3}{c}{\liberolong{}} & \multicolumn{3}{c}{\liberospatial{}}  \\
\cmidrule(lr){3-5}\cmidrule(lr){6-8}

\multirow{3}{*}{\seql{}}    & \bcrnn                    &  0.24 \fs {   0.02 } &  0.28  \fs {   0.01 } &  0.07  \fs {  0.01 } &  0.50 \fs {   0.01 } &  0.61  \fs {   0.01 } &  0.14  \fs {  0.01 } \\
                            & \bct                      &  0.54 \fs {   0.01 } &  0.63  \fs {   0.01 } &  0.15  \fs {  0.00 } &  0.72 \fs {   0.01 } &  0.81  \fs {   0.01 } &  0.20  \fs {  0.01 } \\
                            & \bcvilt                   &  0.44 \fs {   0.04 } &  0.50  \fs {   0.05 } &  0.13  \fs {  0.01 } &  0.63 \fs {   0.02 } &  0.76  \fs {   0.01 } &  0.16  \fs {  0.01 } \\
\cmidrule(lr){2-8}

\multirow{3}{*}{\er{}}      & \bcrnn                    &  0.16 \fs {   0.02 } &  0.16  \fs {   0.02 } &  0.08  \fs {  0.01 } &  0.40 \fs {   0.02 } &  0.29  \fs {   0.02 } &  0.29  \fs {  0.01 } \\
                            & \bct                      &  0.48 \fs {   0.02 } &  0.32  \fs {   0.04 } &  \toptwoboxit 0.32  \fs {  0.01 } & 0.65 \fs {   0.03 } & 0.27  \fs {   0.03 } &  \topthreeboxit 0.56  \fs {  0.01 } \\
                            & \bcvilt                   &  0.38 \fs {   0.05 } & 0.29  \fs {   0.06 } &  \topthreeboxit 0.25  \fs {  0.02 } &  0.63 \fs {   0.01 } &  0.29  \fs {   0.02 } &  0.50  \fs {  0.02 } \\
\cmidrule(lr){2-8}

\multirow{3}{*}{\ewc{}}     & \bcrnn                    &  0.02 \fs {   0.00 } &  0.04  \fs {   0.01 } &  0.00  \fs {  0.00 } &  0.14 \fs {   0.02 } &  0.23  \fs {   0.02 } &  0.03  \fs {  0.00 } \\
                            & \bct                      &  0.13 \fs {   0.02 } &  0.22  \fs {   0.03 } &  0.02  \fs {  0.00 } &  0.23 \fs {   0.01 } &  0.33  \fs {   0.01 } &  0.06  \fs {  0.01 } \\
                            & \bcvilt                   &  0.05 \fs {   0.02 } &  0.09  \fs {   0.03 } &  0.01  \fs {  0.00 } &  0.32 \fs {   0.03 } &  0.48  \fs {   0.03 } &  0.06  \fs {  0.01 } \\
\cmidrule(lr){2-8}

\multirow{3}{*}{\packnet{}} & \bcrnn                    &  0.13 \fs {   0.00 } &  0.21  \fs {   0.01 } &  0.03  \fs {  0.00 } &  0.27 \fs {   0.03 } &  0.38  \fs {   0.03 } &  0.06  \fs {  0.01 } \\
                            & \bct                      &  0.22 \fs {   0.01 } &  0.08  \fs {   0.01 } &  \topthreeboxit 0.25  \fs {  0.00 } &  0.55 \fs {   0.01 } &  0.07  \fs {   0.02 } &  \toponeboxit 0.63  \fs {  0.00 } \\
                            & \bcvilt                   &  0.36 \fs {   0.01 } &  0.14  \fs {   0.01 } &   \toponeboxit 0.34  \fs {  0.01 } & 0.57 \fs {   0.04 } &  0.15  \fs {   0.00 } & \toptwoboxit 0.59  \fs {  0.03 } \\
\cmidrule(lr){2-8}

\multirow{3}{*}{\mtl{}}     & \bcrnn                    &                      &                       &  0.20  \fs {  0.01 }  &                      &                       &  0.61  \fs {  0.00 } \\
                            & \bct                      &                      &                       &  0.48  \fs {  0.01 }  &                      &                       &  0.83  \fs {  0.00 } \\
                            & \bcvilt                   &                      &                       &  0.46  \fs {  0.00 }  &                      &                       &  0.79  \fs {  0.01 } \\
\midrule 
    & & \multicolumn{3}{c}{\liberoobject{}} & \multicolumn{3}{c}{\liberogoal{}}\\
\cmidrule(lr){3-5} \cmidrule(lr){6-8}
\multirow{3}{*}{\seql{}}    & \bcrnn                    &  0.48 \fs {   0.03 } &  0.53  \fs {   0.04 } &  0.15  \fs {  0.01 } &  0.61 \fs {   0.01 } &  0.73  \fs {   0.01 } &  0.16  \fs {  0.00 } \\
                            & \bct                      &  0.78 \fs {   0.04 } &  0.76  \fs {   0.04 } &  0.26  \fs {  0.02 } &  0.77 \fs {   0.01 } &  0.82  \fs {   0.01 } &  0.22  \fs {  0.00 } \\
                            & \bcvilt                   &  0.76 \fs {   0.03 } &  0.73  \fs {   0.03 } &  0.27  \fs {  0.02 } &  0.75 \fs {   0.01 } &  0.85  \fs {   0.01 } &  0.20  \fs {  0.01 } \\
                  \cmidrule(lr){2-8}
\multirow{3}{*}{\er{}}      & \bcrnn                    &  0.30 \fs {   0.01 } &  0.27  \fs {   0.05 } &  0.17  \fs {  0.05 } &  0.41 \fs {   0.00 } &  0.35  \fs {   0.01 } &  0.26  \fs {  0.01 } \\
                            & \bct                      &  0.67 \fs {   0.07 } &  0.43  \fs {   0.04 } &  0.44  \fs {  0.06 } &  0.64 \fs {   0.01 } &  0.34  \fs {   0.02 } &  \topthreeboxit 0.49  \fs {  0.02 } \\
                            & \bcvilt                   &  0.70 \fs {   0.02 } &  0.28  \fs {   0.01 } &  \toptwoboxit 0.57  \fs {  0.01 } &  0.57 \fs {   0.00 } &  0.40  \fs {   0.02 } &  0.38  \fs {  0.01 } \\
                  \cmidrule(lr){2-8}
\multirow{3}{*}{\ewc{}}     & \bcrnn                    &  0.17 \fs {   0.04 } &  0.23  \fs {   0.04 } &  0.06  \fs {  0.01 } &  0.16 \fs {   0.01 } &  0.22  \fs {   0.01 } &  0.06  \fs {  0.01 } \\
                            & \bct                      &  0.56 \fs {   0.03 } &  0.69  \fs {   0.02 } &  0.16  \fs {  0.02 } &  0.32 \fs {   0.02 } &  0.48  \fs {   0.03 } &  0.06  \fs {  0.00 } \\
                            & \bcvilt                   &  0.57 \fs {   0.03 } &  0.64  \fs {   0.03 } &  0.23  \fs {  0.00 } &  0.32 \fs {   0.04 } &  0.45  \fs {   0.04 } &  0.07  \fs {  0.01 } \\
                  \cmidrule(lr){2-8}
\multirow{3}{*}{\packnet{}} & \bcrnn                    &  0.29 \fs {   0.02 } &  0.35  \fs {   0.02 } &  0.13  \fs {  0.01 } &  0.32 \fs {   0.03 } &  0.37  \fs {   0.04 } &  0.11  \fs {  0.01 } \\
                          & \bct                      &  0.60 \fs {   0.07 } & 0.17  \fs {   0.05 } &  \toponeboxit 0.60  \fs {  0.05 } & 0.63 \fs {   0.02 } & 0.06  \fs {   0.01 } & \toptwoboxit 0.75  \fs {  0.01 } \\
                          & \bcvilt                   &  0.58 \fs {   0.03 } &  0.18  \fs {   0.02 } &  \topthreeboxit 0.56  \fs {  0.04 } & 0.69 \fs {   0.02 } &  0.08  \fs {   0.01 } &  \toponeboxit 0.76  \fs {  0.02 } \\
                  \cmidrule(lr){2-8}
\multirow{3}{*}{\mtl{}}   & \bcrnn                    &                      &                       &  0.10  \fs {  0.03 }  &                      &                       &  0.59  \fs {  0.00 } \\
                          & \bct                      &                      &                       &  0.54  \fs {  0.02 }  &                      &                       &  0.80  \fs {  0.01 } \\
                          & \bcvilt                   &                      &                       &  0.78  \fs {  0.02 }  &                      &                       &  0.82  \fs {  0.01 } \\
\bottomrule

    \end{tabular}
    }
    \caption{
We present the full results of all networks and algorithms on all four task suites. For each task suite, we highlight the top three AUC scores among the combinations of the three lifelong learning algorithms and the three neural architectures. The best three results are highlighted in \textbf{\textcolor{top1_boxit_color!90}{magenta}} (the best), \textbf{\textcolor{top2_boxit_color!50}{light magenta}} (the second best), and \textbf{\textcolor{top3_boxit_color!20}{super light magenta}} (the third best), respectively.
    }
    \label{tab:algorithm}
\end{table}

\begin{figure}[t!]
    \centering
    \includegraphics[width=\textwidth]{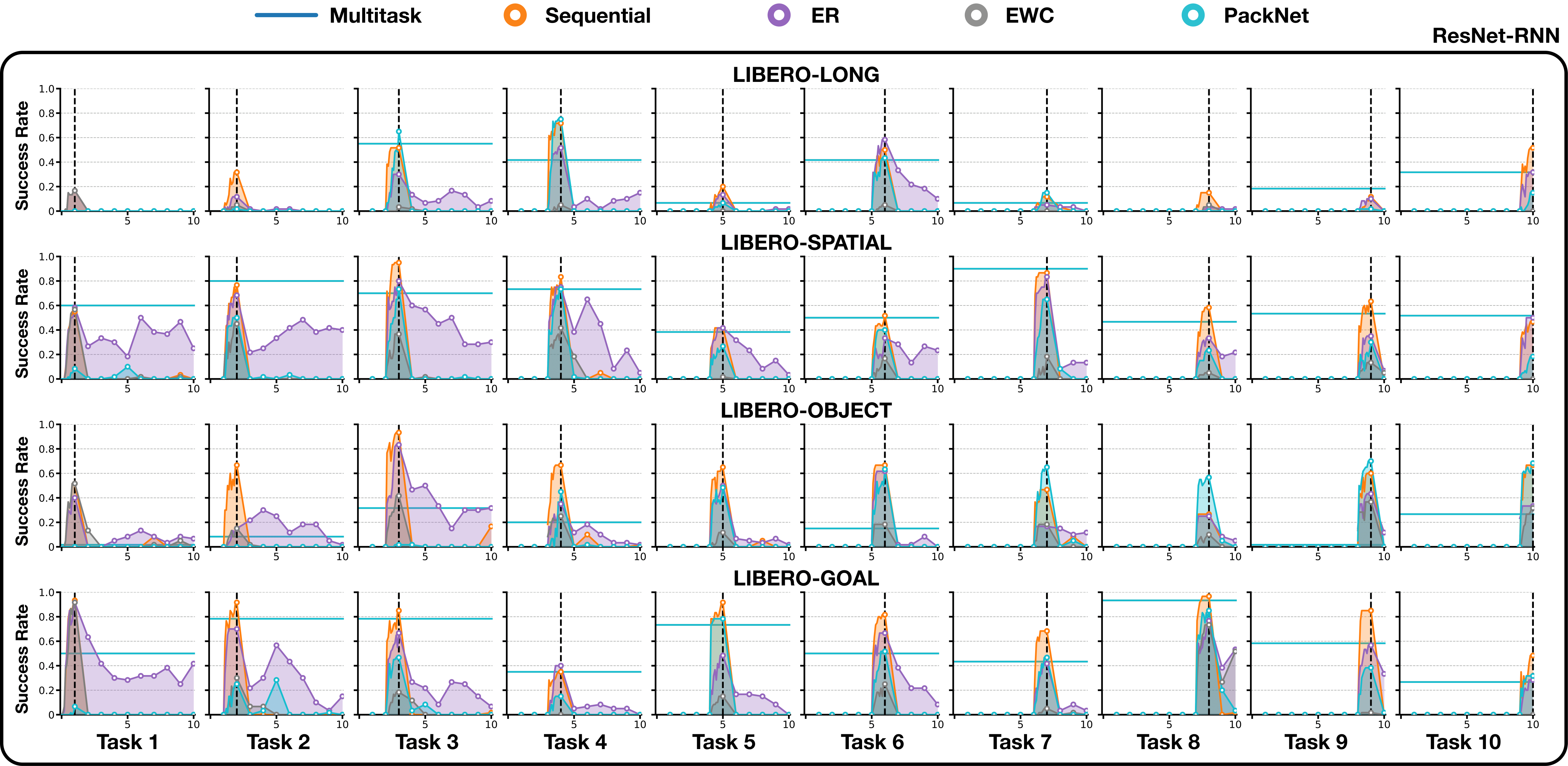}
    \caption{We compare the performance of different algorithms using the \bcrnn{} policy architecture in Figure~\ref{fig:algo_resnet_rnn}. The $y$-axis represents the success rate, and the $x$-axis shows the agent's performance on each of the 10 tasks in a specific task suite over the course of learning. For example, the upper-left plot in the figure displays the agent's performance on the first task as it learns the 10 tasks sequentially.}
    \label{fig:algo_resnet_rnn}
\end{figure}

\begin{figure}[t!]
    \centering
    \includegraphics[width=\textwidth]{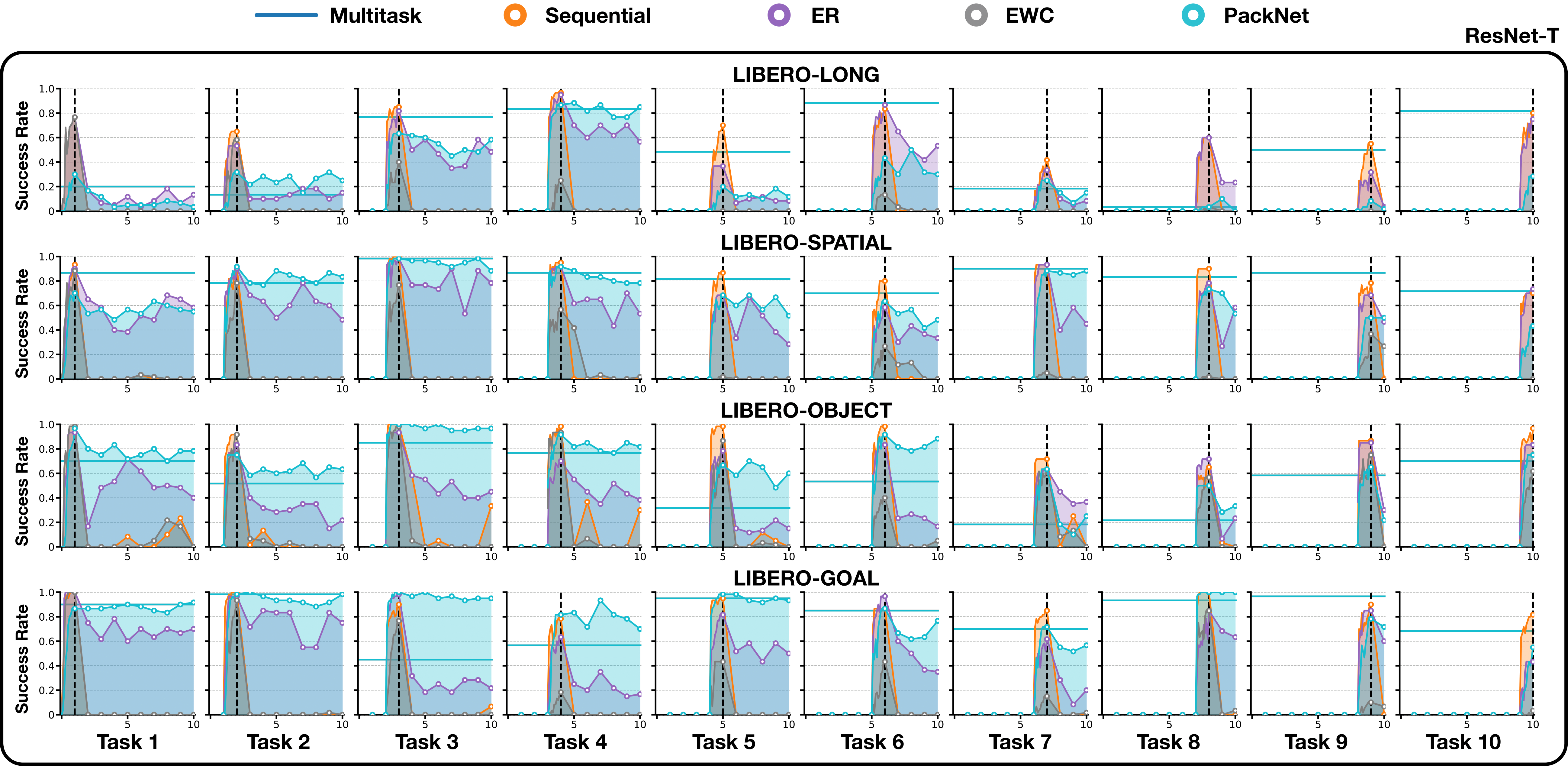}
    \caption{Comparison of different algorithms using the \bct{} policy architecture. The $y$-axis represents the success rate, while the $x$-axis shows the agent's performance on each of the 10 tasks in a given task suite during the course of learning. For example, the plot in the upper-left corner depicts the agent's performance on the first task as it learns the 10 tasks sequentially.}
    \label{fig:algo_resnet_t}
\end{figure}

\begin{figure}[t!]
    \centering
    \includegraphics[width=\textwidth]{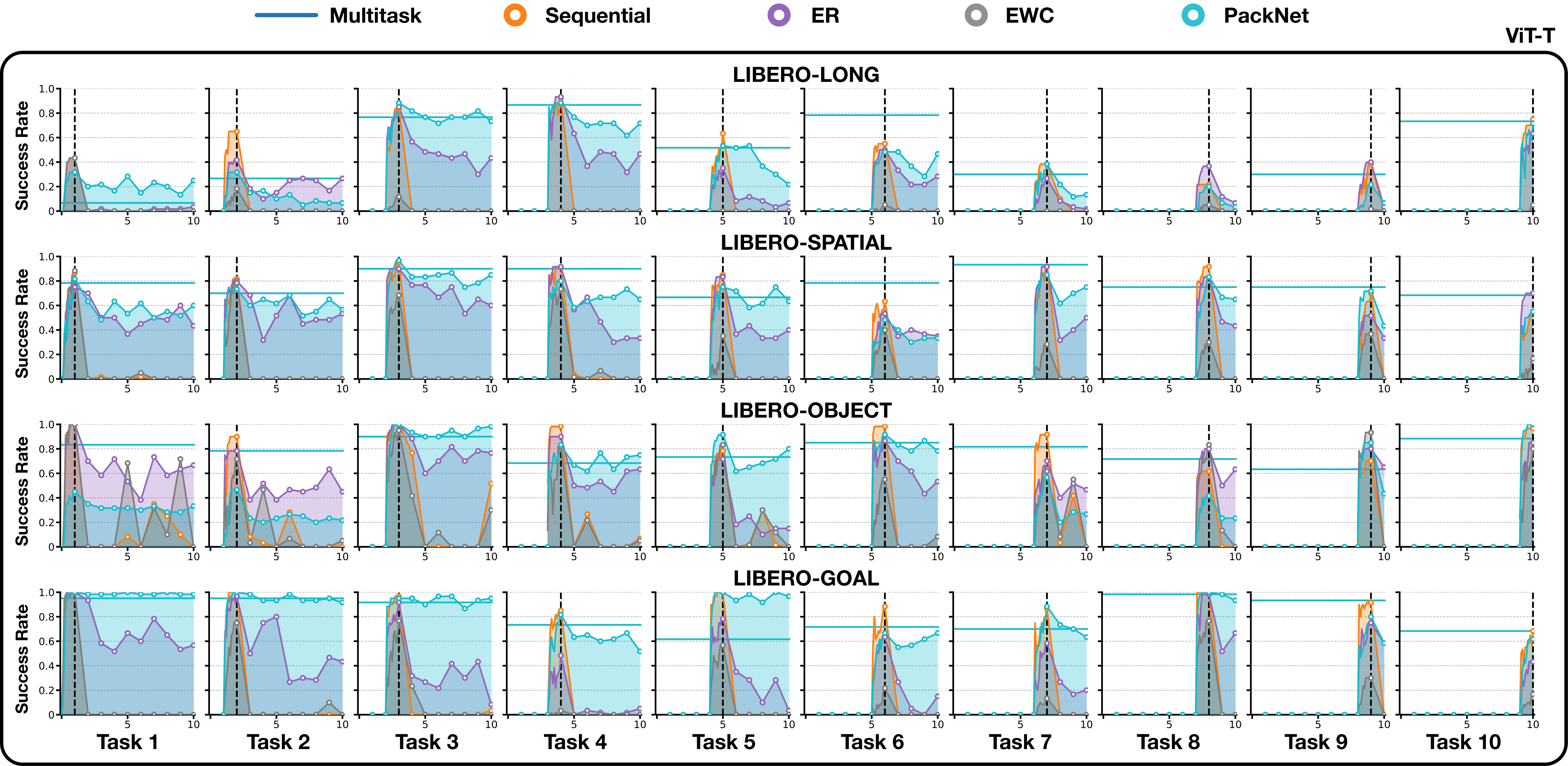}
    \caption{Comparison of different algorithms using the \bcvilt{} policy architecture. The success rate is represented on the $y$-axis, while the $x$-axis shows the agent's performance on the 10 tasks in a given task suite over the course of learning. For instance, the plot in the upper-left corner illustrates the agent's performance on the first task when learning the 10 tasks sequentially. }
    \label{fig:algo_vit_t}
\end{figure}

\begin{figure}[t!]
    \centering
    \includegraphics[width=\textwidth]{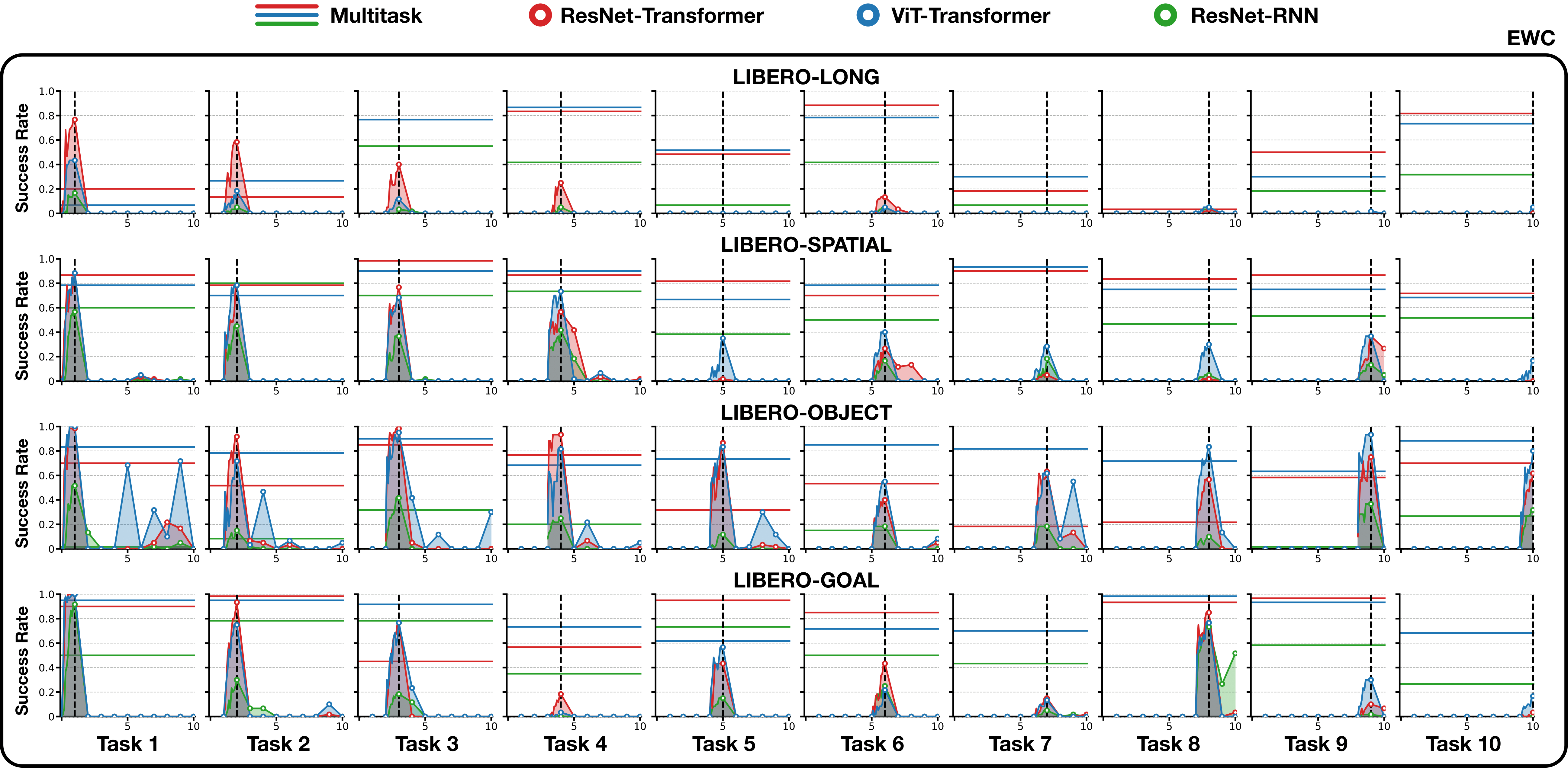}
    \caption{Comparison of different architectures with the \ewc{} algorithm. The $y$-axis is the success rate, while the $x$-axis shows the agent's performance on the 10 tasks in a given task suite over the course of learning. For instance, the upper-left plot shows the agent's performance on the first task when learning the 10 tasks sequentially. }
    \label{fig:arch_ewc}
\end{figure}

\begin{figure}[t!]
    \centering
    \includegraphics[width=\textwidth]{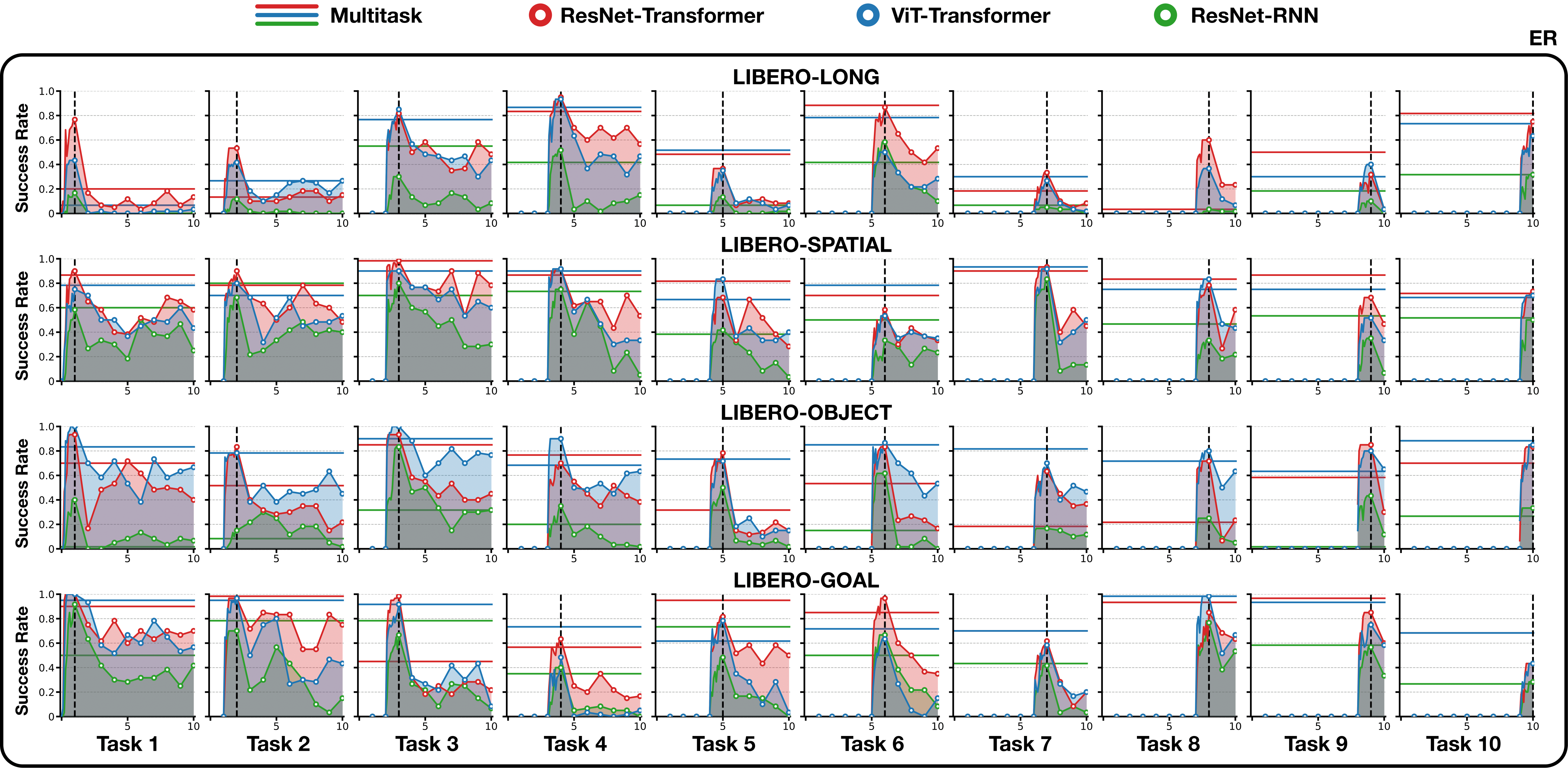}
    \caption{Comparison of different architectures with the \er{} algorithm. The $y$-axis is the success rate, while the $x$-axis shows the agent's performance on the 10 tasks in a given task suite ver the course of learning. For instance, the upper-left plot shows the agent's performance on the first task when learning the 10 tasks sequentially. }
    \label{fig:arch_er}
\end{figure}

\begin{figure}[h!]
    \centering
    \includegraphics[width=\textwidth]{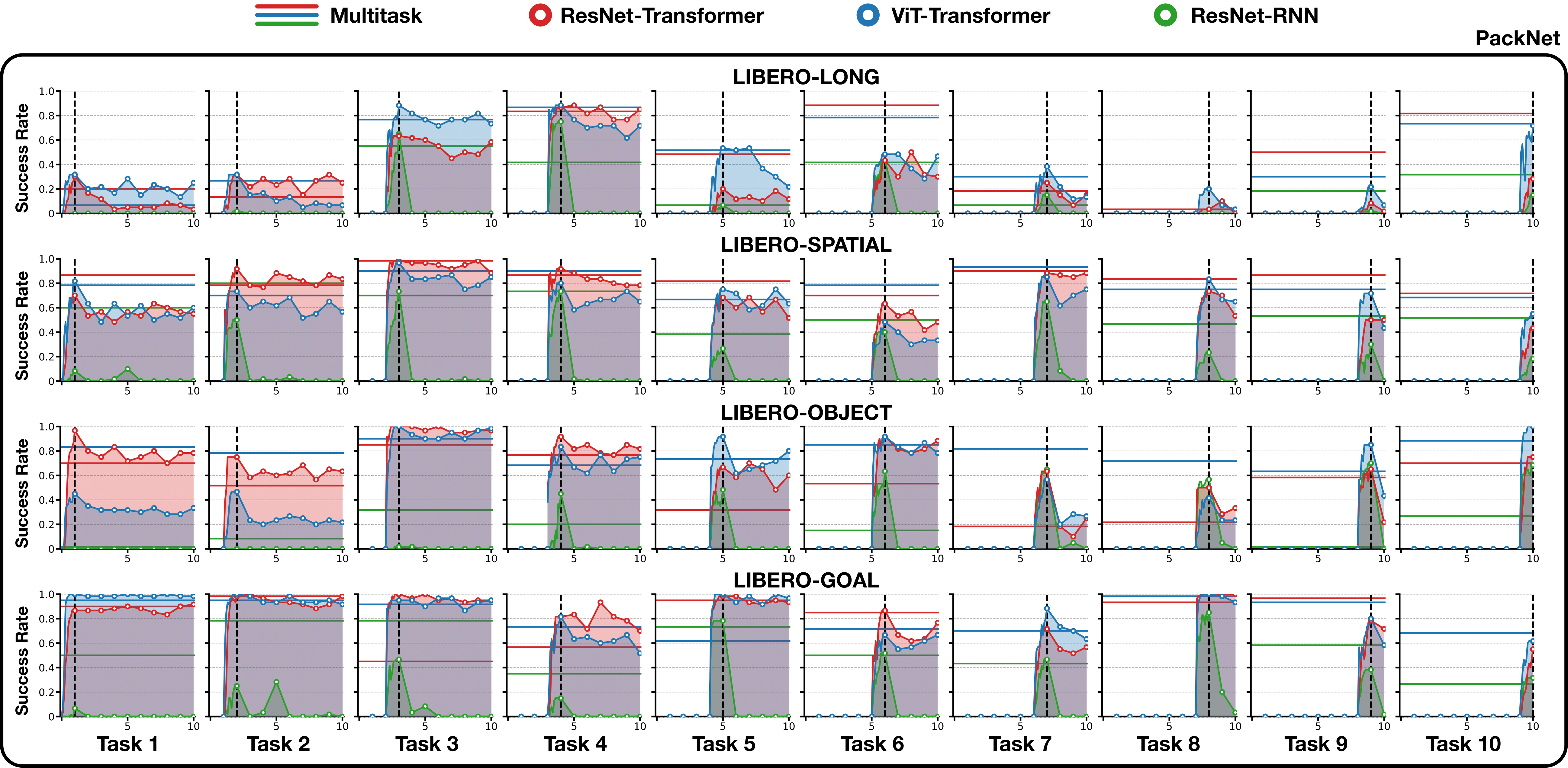}
    \caption{Comparison of different architectures with the \packnet{} algorithm. The $y$-axis is the success rate, while the $x$-axis shows the agent's performance on the 10 tasks in a given task suite over the course of learning. For instance, the upper-left plot shows the agent's performance on the first task when learning the 10 tasks sequentially. }
    \label{fig:arch_packnet}
\end{figure}

To better illustrate the performance of each lifelong learning agent throughout the learning process, we present plots that show how the agent's performance evolves over the stream of tasks. Firstly, we provide plots that compare the performance of the agent using different lifelong learning algorithms while fixing the policy architecture (refer to Figure~\ref{fig:algo_resnet_rnn},\ref{fig:algo_resnet_t}, and \ref{fig:algo_vit_t}). Next, we provide plots that compare the performance of the agent using different policy architectures while fixing the lifelong learning algorithm (refer to Figure\ref{fig:arch_ewc},~\ref{fig:arch_er}, and \ref{fig:arch_packnet})


\clearpage

\clearpage


\clearpage

\subsection{Loss v.s. Success Rates}
\label{appendix:loss-success-rates}
\begin{figure}[h!]
    \centering
    \includegraphics[width=\textwidth]{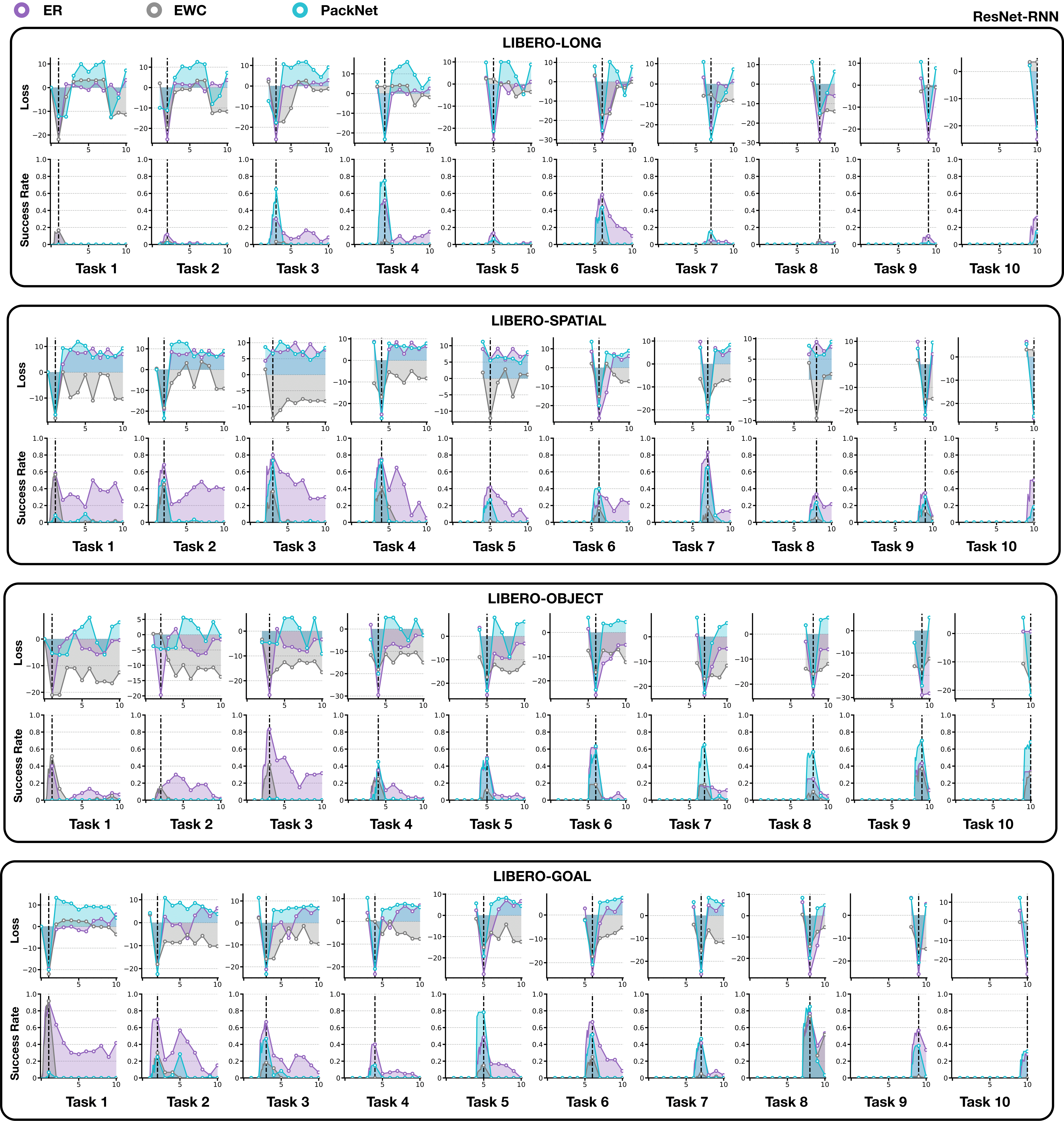}
    \caption{Losses and success rates of \er{} (violet), \ewc{} (grey), and \packnet{} (blue) on four task suites with \bcrnn{} policy. The first (second) row shows the loss (success rate) of the agent on task $i$ throughout the \lldm{} procedure.}
    \label{fig:lsr_resnet_rnn}
\end{figure}

We demonstrate that behavioral cloning loss can be a misleading indicator of task success rate in this section. In supervised learning tasks like image classifications, lower loss often indicates better prediction accuracy. However, this is not, in general, true for decision-making tasks. This is because errors can compound until failures during executing a robot~\citep{ross2011reduction}. Figure~\ref{fig:lsr_resnet_rnn},~\ref{fig:algo_resnet_t} and~\ref{fig:algo_vit_t} plots the training loss and success rates of three lifelong learning methods (\er{}, \ewc{}, and \packnet{}) for comparison. We evaluate the three algorithms on four task suites using three different neural architectures. 

\textcolor{purple}{\textit{Findings:}} We observe that though sometimes \ewc{} has the \textbf{lowest} loss, it did not achieve good success rate. \er{}, on the other hand, can have the highest loss but perform better than \ewc{}. In conclusion, success rates, instead of behavioral cloning loss, should be the right metric to evaluate whether a model checkpoint is good or not.

\clearpage

\clearpage

\begin{figure}[h!]
    \centering
    \includegraphics[width=\textwidth]{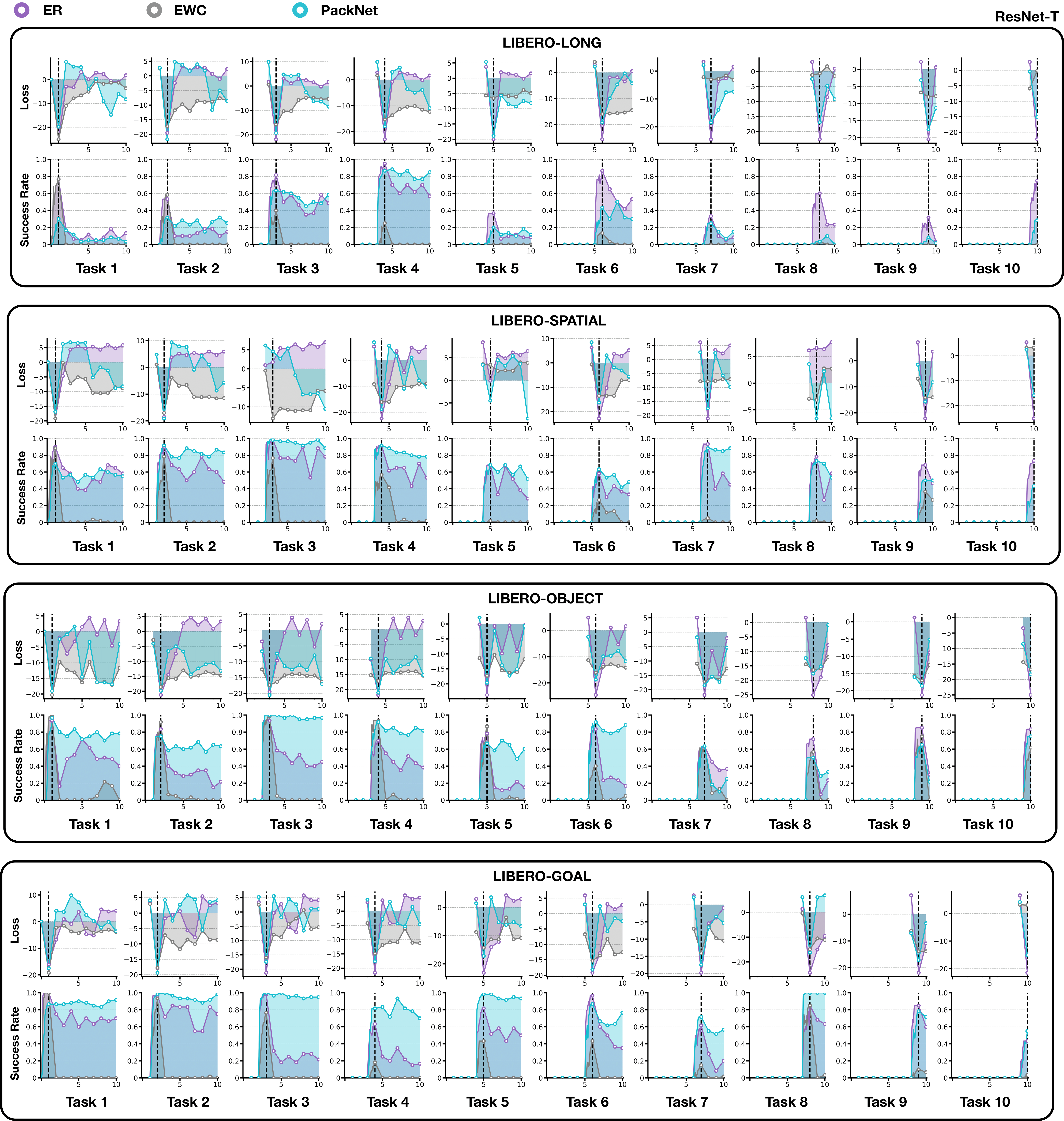}
    \caption{Losses and success rates of \er{} (violet), \ewc{} (grey), and \packnet{} (blue) on four task suites with \bct{} policy. The first (second) row shows the loss (success rate) of the agent on task $i$ throughout the \lldm{} procedure.}
    \label{fig:lsr_resnet_t}
\end{figure}

\clearpage

\begin{figure}[h!]
    \centering
    \includegraphics[width=\textwidth]{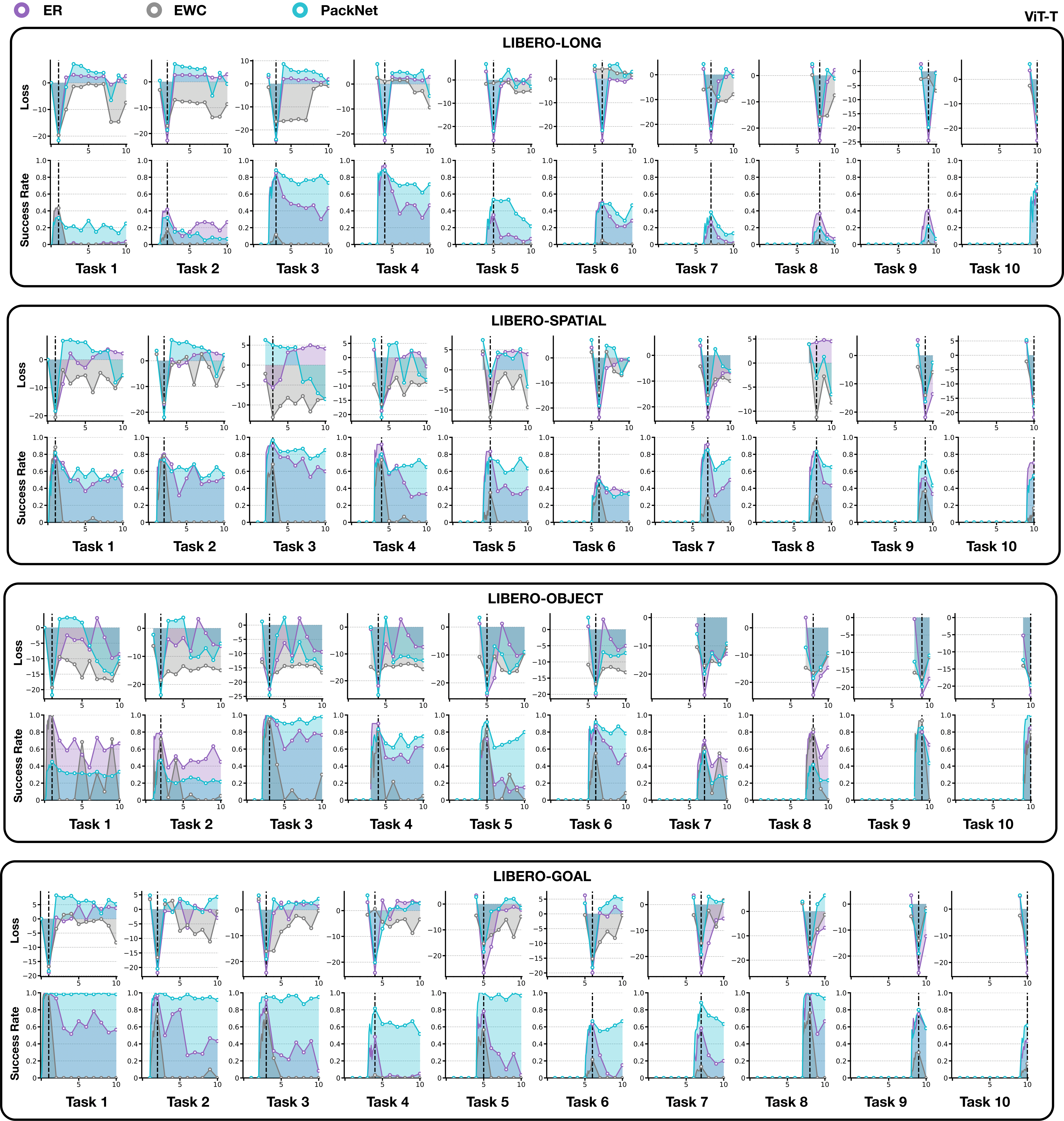}
    \caption{Losses and success rates of \er{} (violet), \ewc{} (grey), and \packnet{} (blue) on four task suites with \bcvilt{} policy. The first (second) row shows the loss (success rate) of the agent on task $i$ throughout the \lldm{} procedure.}
    \label{fig:lsr_vit_t}
\end{figure}

\clearpage
\subsection{Multitask Success Rate}
\label{appendix:multi-succ}

\begin{table}[h!]
    \centering
\resizebox{\textwidth}{!}{
\begin{tabular}{l c c c}
\toprule
Env Name & \bcrnn{} & \bct{} & \bcvilt{} \\ 
\midrule
KITCHEN SCENE10 close the top drawer of the cabinet & 0.45 & 0.45 & 0.6 \\ 
\midrule
KITCHEN SCENE10 close the top drawer of the cabinet and put the black bowl on top of it & 0.1 & 0.1 & 0.0 \\ 
\midrule
KITCHEN SCENE10 put the black bowl in the top drawer of the cabinet & 0.0 & 0.25 & 0.05 \\ 
\midrule
KITCHEN SCENE10 put the butter at the back in the top drawer of the cabinet and close it & 0.15 & 0.0 & 0.0 \\ 
\midrule
KITCHEN SCENE10 put the butter at the front in the top drawer of the cabinet and close it & 0.2 & 0.0 & 0.0 \\ 
\midrule
KITCHEN SCENE10 put the chocolate pudding in the top drawer of the cabinet and close it & 0.25 & 0.0 & 0.0 \\ 
\midrule
KITCHEN SCENE1 open the bottom drawer of the cabinet & 0.05 & 0.0 & 0.15 \\ 
\midrule
KITCHEN SCENE1 open the top drawer of the cabinet & 0.3 & 0.45 & 0.25 \\ 
\midrule
KITCHEN SCENE1 open the top drawer of the cabinet and put the bowl in it & 0.05 & 0.0 & 0.0 \\ 
\midrule
KITCHEN SCENE1 put the black bowl on the plate & 0.35 & 0.3 & 0.0 \\ 
\midrule
KITCHEN SCENE1 put the black bowl on top of the cabinet & 0.2 & 0.7 & 0.15 \\ 
\midrule
KITCHEN SCENE2 open the top drawer of the cabinet & 0.45 & 0.4 & 0.65 \\ 
\midrule
KITCHEN SCENE2 put the black bowl at the back on the plate & 0.2 & 0.05 & 0.35 \\ 
\midrule
KITCHEN SCENE2 put the black bowl at the front on the plate & 0.1 & 0.35 & 0.0 \\ 
\midrule
KITCHEN SCENE2 put the middle black bowl on the plate & 0.35 & 0.1 & 0.0 \\ 
\midrule
KITCHEN SCENE2 put the middle black bowl on top of the cabinet & 0.5 & 0.1 & 0.6 \\ 
\midrule
KITCHEN SCENE2 stack the black bowl at the front on the black bowl in the middle & 0.25 & 0.05 & 0.25 \\ 
\midrule
KITCHEN SCENE2 stack the middle black bowl on the back black bowl & 0.05 & 0.05 & 0.0 \\ 
\midrule
KITCHEN SCENE3 put the frying pan on the stove & 0.4 & 0.3 & 0.35 \\ 
\midrule
KITCHEN SCENE3 put the moka pot on the stove & 0.15 & 0.0 & 0.4 \\ 
\midrule
KITCHEN SCENE3 turn on the stove & 0.6 & 0.7 & 1.0 \\ 
\midrule
KITCHEN SCENE3 turn on the stove and put the frying pan on it & 0.15 & 0.0 & 0.35 \\ 
\midrule
KITCHEN SCENE4 close the bottom drawer of the cabinet & 0.75 & 0.4 & 0.55 \\ 
\midrule
KITCHEN SCENE4 close the bottom drawer of the cabinet and open the top drawer & 0.2 & 0.0 & 0.05 \\ 
\midrule
KITCHEN SCENE4 put the black bowl in the bottom drawer of the cabinet & 0.25 & 0.3 & 0.25 \\ 
\midrule
KITCHEN SCENE4 put the black bowl on top of the cabinet & 0.8 & 0.6 & 0.85 \\ 
\midrule
KITCHEN SCENE4 put the wine bottle in the bottom drawer of the cabinet & 0.0 & 0.0 & 0.35 \\ 
\midrule
KITCHEN SCENE4 put the wine bottle on the wine rack & 0.05 & 0.0 & 0.2 \\ 
\midrule
KITCHEN SCENE5 close the top drawer of the cabinet & 0.05 & 0.8 & 0.9 \\ 
\midrule
KITCHEN SCENE5 put the black bowl in the top drawer of the cabinet & 0.2 & 0.1 & 0.15 \\ 
\midrule
KITCHEN SCENE5 put the black bowl on the plate & 0.05 & 0.05 & 0.1 \\ 
\midrule
KITCHEN SCENE5 put the black bowl on top of the cabinet & 0.5 & 0.25 & 0.2 \\ 
\midrule
KITCHEN SCENE5 put the ketchup in the top drawer of the cabinet & 0.0 & 0.0 & 0.05 \\ 
\midrule
KITCHEN SCENE6 close the microwave & 0.1 & 0.1 & 0.2 \\ 
\midrule
KITCHEN SCENE6 put the yellow and white mug to the front of the white mug & 0.2 & 0.05 & 0.1 \\ 
\midrule
KITCHEN SCENE7 open the microwave & 0.7 & 0.1 & 0.45 \\ 
\midrule
KITCHEN SCENE7 put the white bowl on the plate & 0.05 & 0.0 & 0.05 \\ 
\midrule
KITCHEN SCENE7 put the white bowl to the right of the plate & 0.05 & 0.05 & 0.15 \\ 
\midrule
KITCHEN SCENE8 put the right moka pot on the stove & 0.15 & 0.0 & 0.1 \\ 
\midrule
KITCHEN SCENE8 turn off the stove & 0.2 & 0.25 & 0.75 \\ 
\midrule
KITCHEN SCENE9 put the frying pan on the cabinet shelf & 0.45 & 0.15 & 0.05 \\ 
\midrule
KITCHEN SCENE9 put the frying pan on top of the cabinet & 0.25 & 0.4 & 0.3 \\ 
\midrule
KITCHEN SCENE9 put the frying pan under the cabinet shelf & 0.15 & 0.45 & 0.15 \\ 
\midrule
KITCHEN SCENE9 put the white bowl on top of the cabinet & 0.1 & 0.1 & 0.15 \\ 
\midrule
KITCHEN SCENE9 turn on the stove & 0.5 & 0.4 & 0.95 \\ 
\midrule
KITCHEN SCENE9 turn on the stove and put the frying pan on it & 0.0 & 0.0 & 0.0 \\ 
\bottomrule
\end{tabular}
}
\end{table}

\clearpage

\begin{table}[h!]
    \centering
\resizebox{\textwidth}{!}{
\begin{tabular}{l c c c}
\toprule
Env Name & \bcrnn{} & \bct{} & \bcvilt{} \\ 
\midrule
LIVING ROOM SCENE1 pick up the alphabet soup and put it in the basket & 0.0 & 0.0 & 0.0 \\ 
\midrule
LIVING ROOM SCENE1 pick up the cream cheese box and put it in the basket & 0.0 & 0.0 & 0.0 \\ 
\midrule
LIVING ROOM SCENE1 pick up the ketchup and put it in the basket & 0.0 & 0.0 & 0.0 \\ 
\midrule
LIVING ROOM SCENE1 pick up the tomato sauce and put it in the basket & 0.0 & 0.0 & 0.0 \\ 
\midrule
LIVING ROOM SCENE2 pick up the alphabet soup and put it in the basket & 0.0 & 0.0 & 0.0 \\ 
\midrule
LIVING ROOM SCENE2 pick up the butter and put it in the basket & 0.0 & 0.0 & 0.0 \\ 
\midrule
LIVING ROOM SCENE2 pick up the milk and put it in the basket & 0.0 & 0.05 & 0.0 \\ 
\midrule
LIVING ROOM SCENE2 pick up the orange juice and put it in the basket & 0.0 & 0.0 & 0.0 \\ 
\midrule
LIVING ROOM SCENE2 pick up the tomato sauce and put it in the basket & 0.0 & 0.05 & 0.0 \\ 
\midrule
LIVING ROOM SCENE3 pick up the alphabet soup and put it in the tray & 0.0 & 0.05 & 0.0 \\ 
\midrule
LIVING ROOM SCENE3 pick up the butter and put it in the tray & 0.0 & 0.3 & 0.0 \\ 
\midrule
LIVING ROOM SCENE3 pick up the cream cheese and put it in the tray & 0.0 & 0.25 & 0.0 \\ 
\midrule
LIVING ROOM SCENE3 pick up the ketchup and put it in the tray & 0.0 & 0.0 & 0.0 \\ 
\midrule
LIVING ROOM SCENE3 pick up the tomato sauce and put it in the tray & 0.0 & 0.25 & 0.0 \\ 
\midrule
LIVING ROOM SCENE4 pick up the black bowl on the left and put it in the tray & 0.0 & 0.4 & 0.2 \\ 
\midrule
LIVING ROOM SCENE4 pick up the chocolate pudding and put it in the tray & 0.0 & 0.2 & 0.25 \\ 
\midrule
LIVING ROOM SCENE4 pick up the salad dressing and put it in the tray & 0.0 & 0.0 & 0.1 \\ 
\midrule
LIVING ROOM SCENE4 stack the left bowl on the right bowl and place them in the tray & 0.0 & 0.0 & 0.0 \\ 
\midrule
LIVING ROOM SCENE4 stack the right bowl on the left bowl and place them in the tray & 0.0 & 0.0 & 0.05 \\ 
\midrule
LIVING ROOM SCENE5 put the red mug on the left plate & 0.0 & 0.0 & 0.05 \\ 
\midrule
LIVING ROOM SCENE5 put the red mug on the right plate & 0.15 & 0.0 & 0.0 \\ 
\midrule
LIVING ROOM SCENE5 put the white mug on the left plate & 0.1 & 0.15 & 0.05 \\ 
\midrule
LIVING ROOM SCENE5 put the yellow and white mug on the right plate & 0.35 & 0.05 & 0.05 \\ 
\midrule
LIVING ROOM SCENE6 put the chocolate pudding to the left of the plate & 0.1 & 0.65 & 0.0 \\ 
\midrule
LIVING ROOM SCENE6 put the chocolate pudding to the right of the plate & 0.05 & 0.55 & 0.0 \\ 
\midrule
LIVING ROOM SCENE6 put the red mug on the plate & 0.0 & 0.2 & 0.0 \\ 
\midrule
LIVING ROOM SCENE6 put the white mug on the plate & 0.0 & 0.2 & 0.0 \\ 
\midrule
STUDY SCENE1 pick up the book and place it in the front compartment of the caddy & 0.0 & 0.0 & 0.05 \\ 
\midrule
STUDY SCENE1 pick up the book and place it in the left compartment of the caddy & 0.2 & 0.05 & 0.0 \\ 
\midrule
STUDY SCENE1 pick up the book and place it in the right compartment of the caddy & 0.0 & 0.1 & 0.0 \\ 
\midrule
STUDY SCENE1 pick up the yellow and white mug and place it to the right of the caddy & 0.0 & 0.3 & 0.35 \\ 
\midrule
STUDY SCENE2 pick up the book and place it in the back compartment of the caddy & 0.35 & 0.3 & 0.0 \\ 
\midrule
STUDY SCENE2 pick up the book and place it in the front compartment of the caddy & 0.25 & 0.1 & 0.0 \\ 
\midrule
STUDY SCENE2 pick up the book and place it in the left compartment of the caddy & 0.25 & 0.45 & 0.05 \\ 
\midrule
STUDY SCENE2 pick up the book and place it in the right compartment of the caddy & 0.0 & 0.2 & 0.0 \\ 
\midrule
STUDY SCENE3 pick up the book and place it in the front compartment of the caddy & 0.2 & 0.0 & 0.0 \\ 
\midrule
STUDY SCENE3 pick up the book and place it in the left compartment of the caddy & 0.4 & 0.45 & 0.15 \\ 
\midrule
STUDY SCENE3 pick up the book and place it in the right compartment of the caddy & 0.0 & 0.05 & 0.05 \\ 
\midrule
STUDY SCENE3 pick up the red mug and place it to the right of the caddy & 0.0 & 0.2 & 0.05 \\ 
\midrule
STUDY SCENE3 pick up the white mug and place it to the right of the caddy & 0.0 & 0.0 & 0.05 \\ 
\midrule
STUDY SCENE4 pick up the book in the middle and place it on the cabinet shelf & 0.05 & 0.2 & 0.05 \\ 
\midrule
STUDY SCENE4 pick up the book on the left and place it on top of the shelf & 0.2 & 0.1 & 0.15 \\ 
\midrule
STUDY SCENE4 pick up the book on the right and place it on the cabinet shelf & 0.0 & 0.25 & 0.05 \\ 
\midrule
STUDY SCENE4 pick up the book on the right and place it under the cabinet shelf & 0.15 & 0.1 & 0.05 \\ 
\bottomrule
\end{tabular}
}
\end{table}

\clearpage


\subsection{Attention Visualization}
\label{appendix:attention}
It is also important to visualize the behavior of the robot and its attention maps during the completion of tasks in the lifelong learning process to give us intuition and qualitative feedback on the performance of different algorithms and architectures. We visualize the attention maps of learned policies with~\citet{Greydanus2017VisualizingAU} and compare them in different studies as in \ref{experiment} to see if the robot correctly pays attention to the right regions of interest in each task.

\paragraph{Perturbation-based attention visualization:}
We use a perturbation-based method~\cite{Greydanus2017VisualizingAU} to extract attention maps from agents. Given an input image $I$, the method applies a Gaussian filter to a pixel location $(i,j)$ to blur the image partially, and produces the perturbed image $\Phi(I, i, j)$. Denote the learned policy as $\pi$ and the inputs to the spatial module (e.g., the last latent representation of resnet or ViT encoder) $\pi_u(I)$ for image $I$. Then we define the saliency score as the Euclidean distance between the latent representations of the original and the blurred images: 
\begin{equation}
    S_{\pi}(i,j) = \frac{1}{2} \bigg|\bigg|\pi_u(I) - \pi_u(\Phi(I, i, j))\bigg|\bigg|^2.
\end{equation}
Intuitively, $S_{\pi}(i,j)$ describes \emph{how much removing information from the region around location $(i, j)$ changes the policy}. In other words, a large $S_{\pi}(i,j)$ indicates that the information around pixel $(i,j)$ is important for the learning agent's decision-making. Instead of calculating the score for every pixel, \cite{Greydanus2017VisualizingAU} found that computing a saliency score for pixel $i$ mod 5 and $j$ mod 5 produced good saliency maps at lower computational costs for Atari games. The final saliency map $P$ is normalized as $P(i,j) = \frac{S_\pi(i,j)}{\sum_{i,j}S_\pi(i,j)}$.

We provide the visualization and our analysis on the following pages.

\clearpage

\textbf{\large Different Task Suites} \\

\begin{figure*}[h!]
    \centering
    \includegraphics[width=0.9\textwidth]{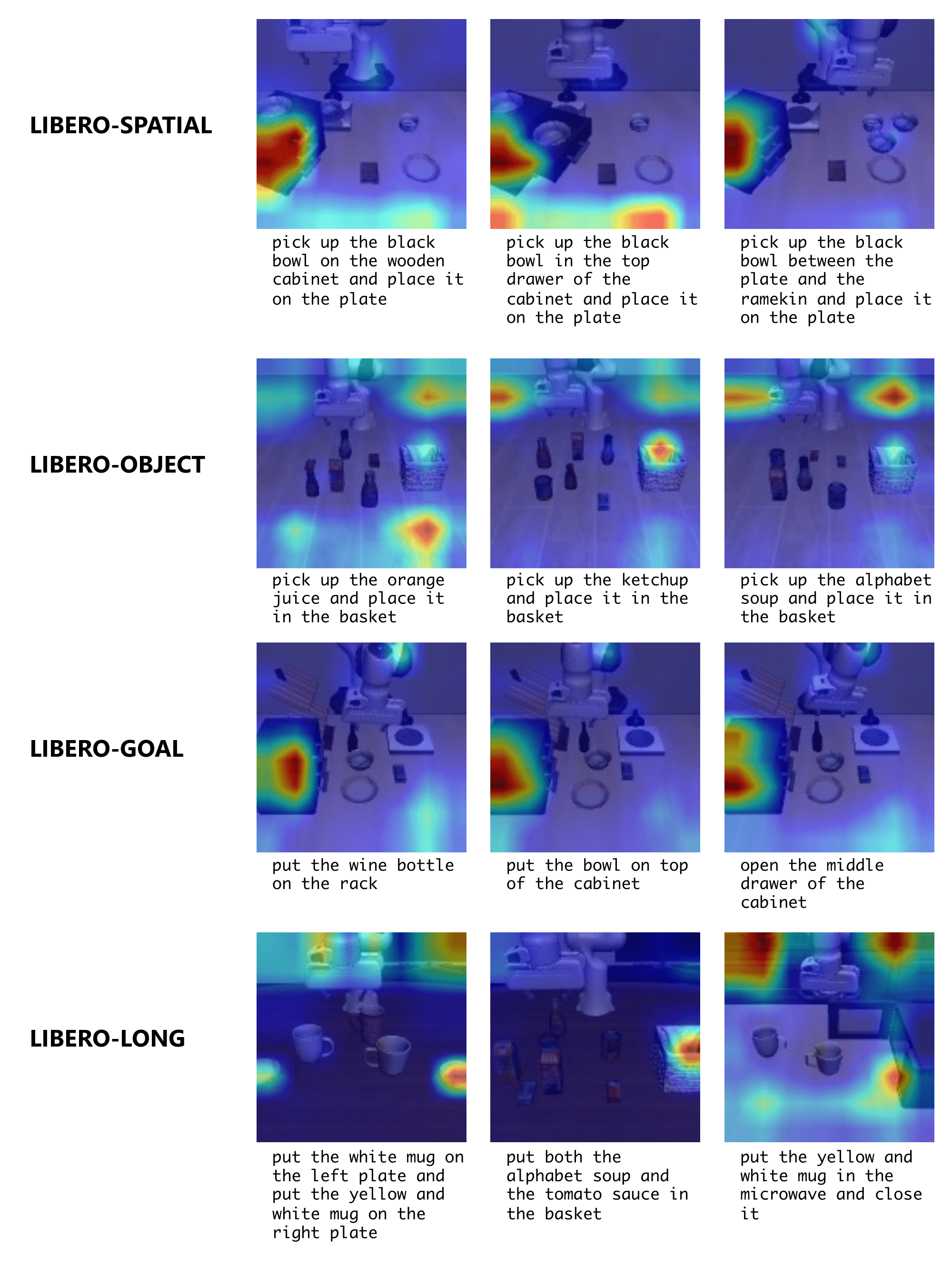}
    \caption{Attention map comparison among different task suites with \er{} and \bct{}. Each row corresponds to a task suite.}
    \label{fig:t1-attention}
\end{figure*}

\textcolor{purple}{\textit{Findings:}}
Figure \ref{fig:t1-attention} shows attention visualization for 12 tasks across 4 task suites (e.g., 3 tasks per suite). We observe that: 
\begin{enumerate}
    \item policies pay more attention to the robot arm and the target placement area than the target object. 
    \item sometimes the policy pays attention to task-irrelevant areas, such as the blank area on the table.
\end{enumerate}
These observations demonstrate that the learned policy use perceptual data for decision-making in a very different way from how humans do. The robot policies tends to spuriously correlate task-irrelevant features with actions, a major reason why the policies overfit to the tasks and do not generalize well across tasks.


\clearpage

\textbf{\large The Same Task over the Course of Lifelong Learning}\\

\begin{figure*}[h!]
    \centering
    \includegraphics[width=0.9\textwidth]{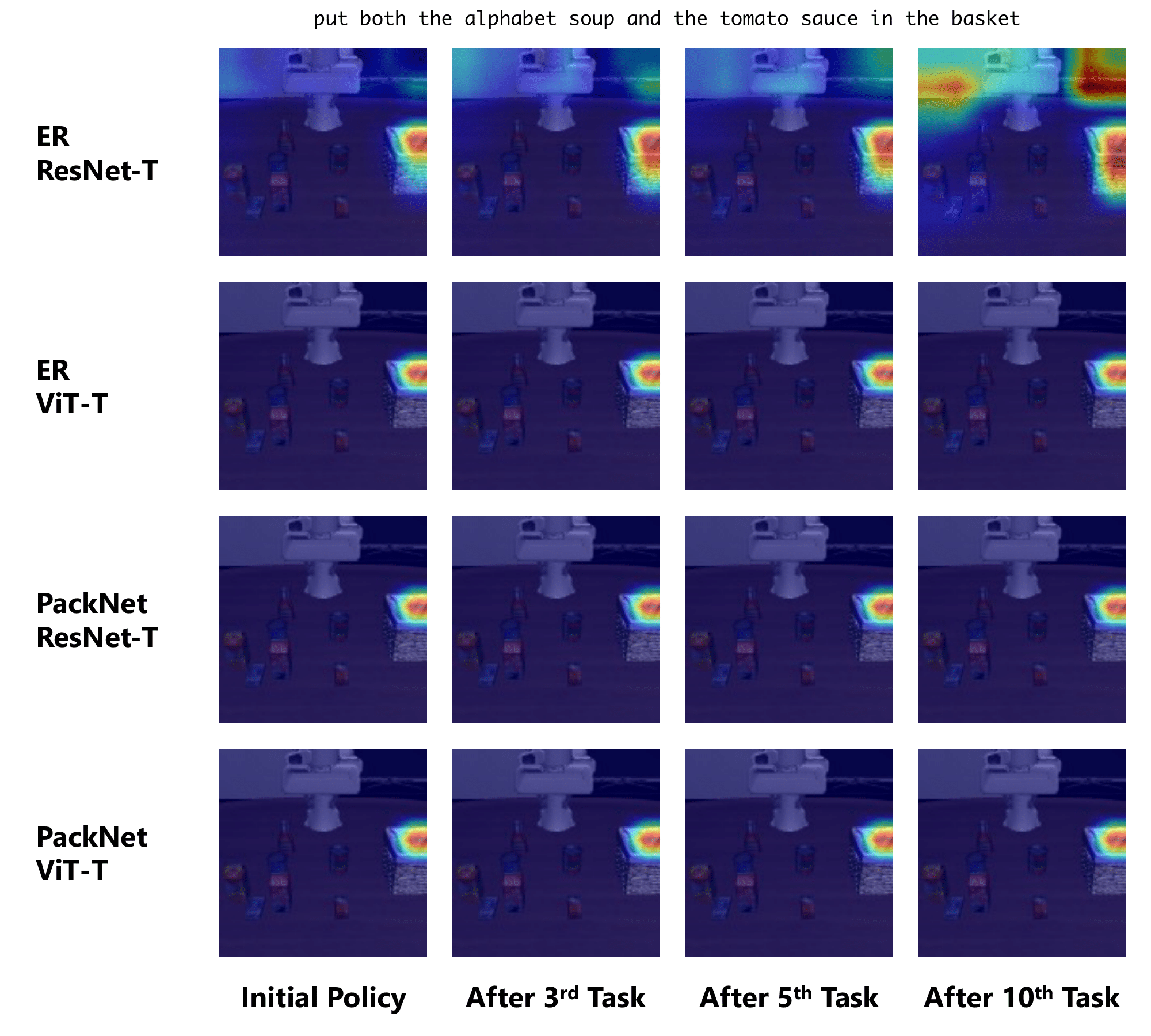}
    \caption{Attention map of the same state of the task \textit{put both the alphabet soup and the tomato sauce in the basket} from \liberolong{} during lifelong learning. Each row visualizes how the attention maps change on the first task with one of the LL algorithms (\er{} and \packnet{}) and one of the neural architectures (\bct{} and \bcvilt{}). Initial policy is the policy that is trained on the first task. And all the following attention maps correspond to policies after training on the third, fifth, and the tenth tasks. 
    }
    \label{fig:lifelong-progress-attention}
\end{figure*}

\textcolor{purple}{\textit{Findings:}}
Figure \ref{fig:lifelong-progress-attention} shows attention visualizations from policies trained with \er{} and \packnet{} using the architectures \bct{} and \bcvilt{} respectively. We observe that:
\begin{enumerate}
    \item The ViT visual encoder's attention is more consistent over time, while the ResNet encoder's attention map gradually dilutes.
    \item PackNet, as it splits the model capacity for different tasks, shows a more consistent attention map over the course of learning.
\end{enumerate}

\clearpage

\textbf{\large Different Lifelong Learning Algorithms} \\

\begin{figure*}[h!]
\centering
\includegraphics[width=0.8\textwidth]{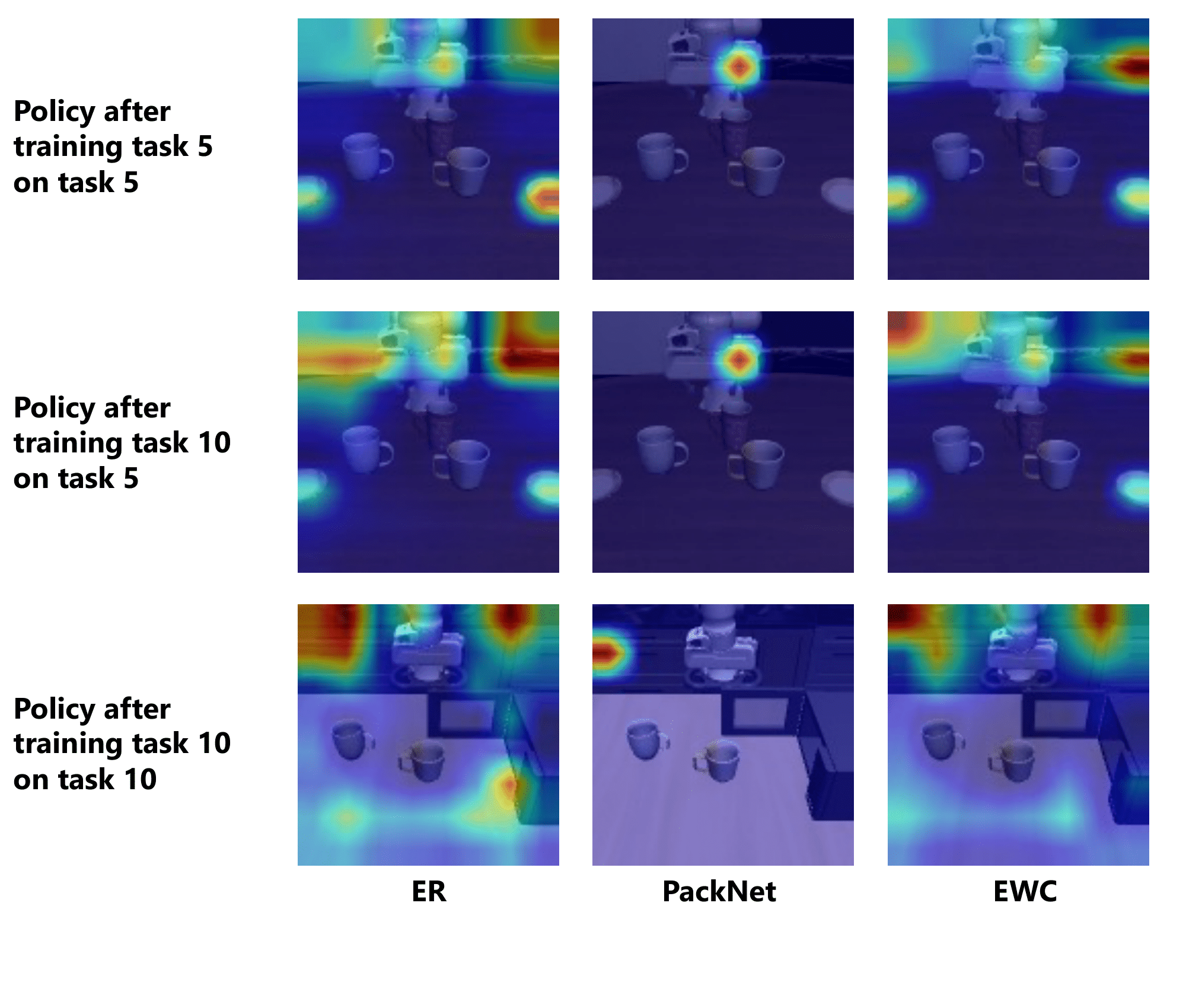}
\caption{Comparison of attention maps of different lifelong learning algorithms with \bct{} on \liberolong{}. Each row shows the same state of a task with different neural architectures. ``Task 5'' refers to the task \textit{put the white mug on the left plate and put the yellow and white mug on the right plate}. ``Task 10'' refers to the task \textit{put the yellow and white mug in the microwave and close it}. The second row shows the policy that is trained on task 10 and gets evaluated on task 5, showing the attention map differences in backward transfer.
}
\label{fig:attn-algo}
\end{figure*}

\textcolor{purple}{\textit{Findings:}}
Figure \ref{fig:attn-algo} shows the attention visualization of three lifelong learning algorithms on \liberolong{} with \bct{} on two tasks (task 5 and task 10). The first and third rows show the attention of the policy on the same task it has just learned. While the second row shows the attention of the policy on the task it learned in the past. We observe that:

\begin{enumerate}
    \item \packnet{} shows more concentrated attention compared against \er{} and \ewc{} (usually just a single mode).
    \item \er{} shares similar attention map with \ewc{}, but \ewc{} performs much worse than \er{}. Therefore, attention can only assist the analysis but cannot be treated as a criterion for performance prediction.
\end{enumerate}


\clearpage 

\textbf{\large Different Neural Architectures}\\

\begin{figure*}[h!]
\centering
\includegraphics[width=0.8\textwidth]{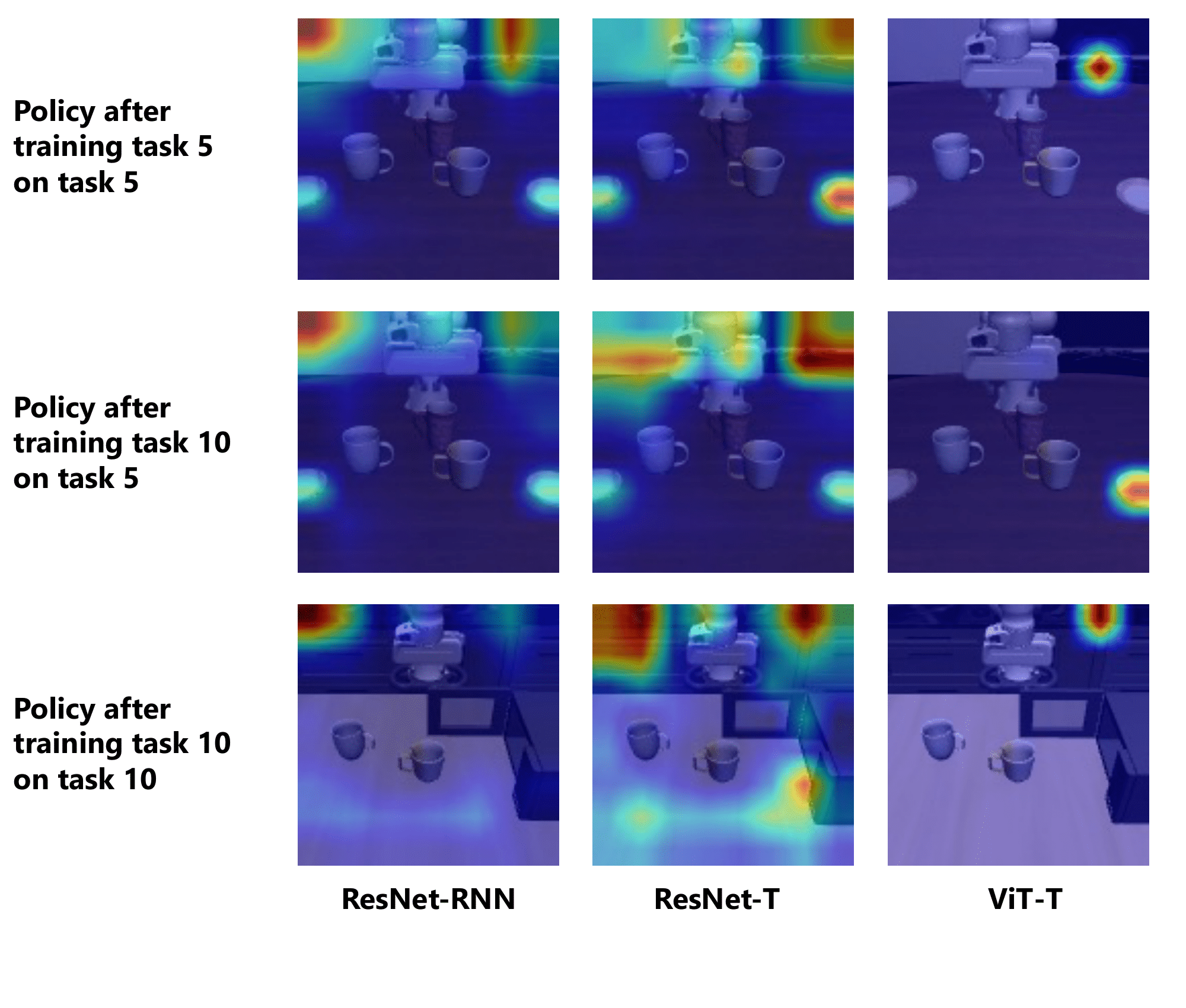}
\caption{Comparison of attention maps of different neural architectures with \er{} on \liberolong{}. Each row shows the same state of a task with different neural architectures. ``Task 5'' refers to the task \textit{put the white mug on the left plate and put the yellow and white mug on the right plate}. ``Task 10'' refers to the task \textit{put the yellow and white mug in the microwave and close it}. The second row shows the policy that is trained on task 10 and gets evaluated on task 5, showing the attention map differences in backward transfer.
}
\label{fig:attn-architecture}
\end{figure*}

\textcolor{purple}{\textit{Findings:}}
Figure \ref{fig:attn-architecture} shows attention map comparisons of the three neural architectures on \liberolong{} with \er{} on two tasks (task 5 and task 10). We observe that:
\begin{enumerate}
    \item ViT has more concentrated attention than policies using ResNet.
    \item When ResNet forgets, the attention is changing smoothly (more diluted). But for ViT, when it forgets, the attention can completely shift to a different location.
    \item When ResNet is combined with LSTM or a temporal transformer, the attention hints at the "course of future trajectory". But we do not observe that when ViT is used as the encoder.
\end{enumerate}

\clearpage

\textbf{\large Different Task Ordering} \\

\begin{figure*}[h!]
    \centering
    \includegraphics[width=0.8\textwidth]{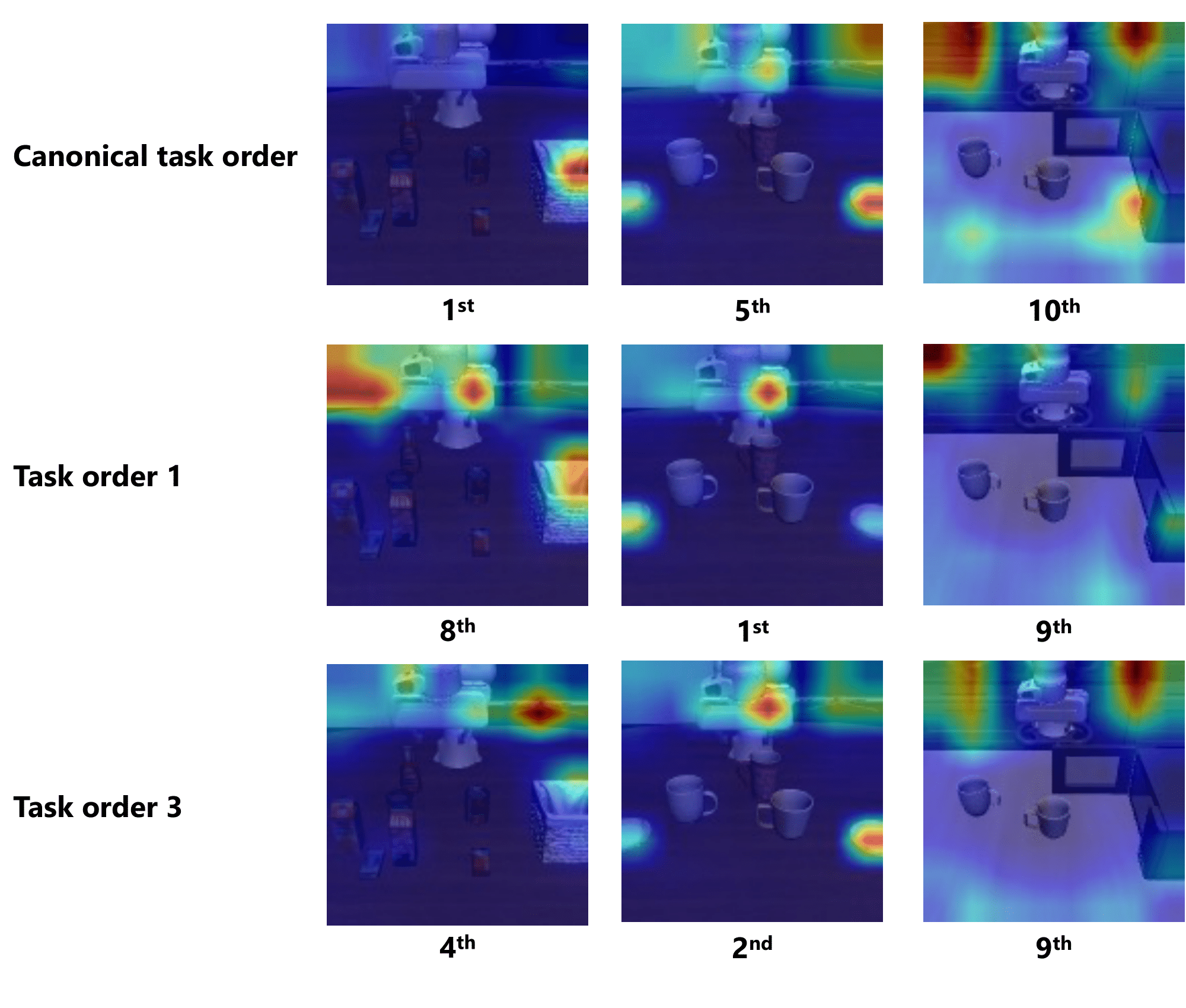}
    \caption{Attention map comparison among different orderings with \er{} and \bct{} on three selected tasks from \liberolong{}: \textit{put both the alphabet soup and the tomato sauce in the basket}, \textit{put the white mug on the left plate and put the yellow and white mug on the right plate}, and \textit{put the yellow and white mug in the microwave and close it}. Each row corresponds to a specific sequence of task ordering, and the caption of each attention map indicates the order of the task in that sequence.}
    \label{fig:t4-attention}
\end{figure*}

\textcolor{purple}{\textit{Findings:}}
Figure \ref{fig:t4-attention} shows attention map comparisons of three different task orderings. We show two immediately learned tasks from \liberolong{} trained with \er{} and \bct{}. We observe that:
\begin{enumerate}
    \item As expected, learning the same task at different positions in the task stream results in different attention visualization.
    \item There seems to be a trend that the policy has a more spread-out attention when it learns on tasks that are later in the sequence.
\end{enumerate}


\clearpage 
\textbf{\large With or Without Pretraining} \\

\begin{figure*}[h!]
    \centering
    \includegraphics[width=0.85\textwidth]{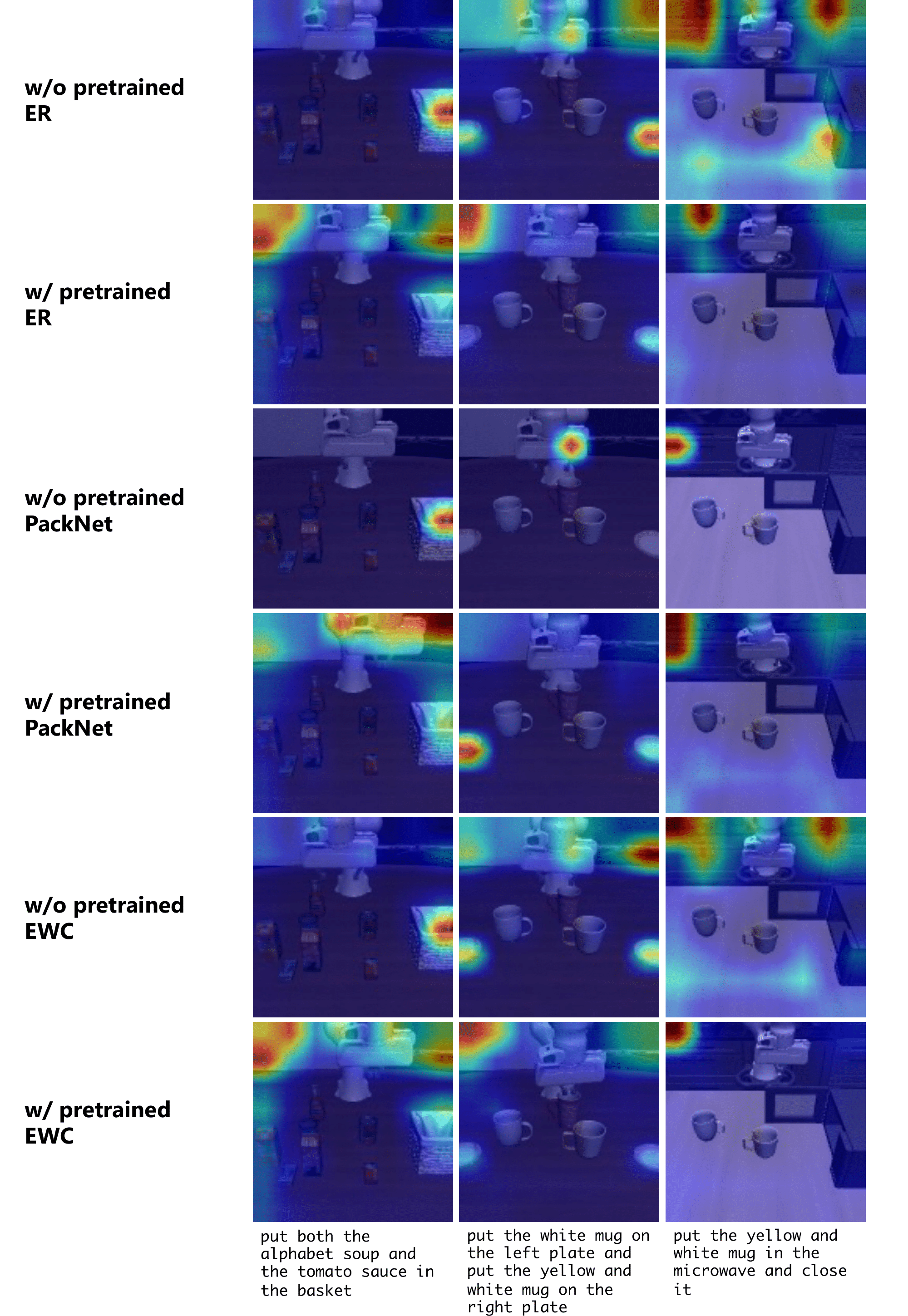}
    \caption{Attention map comparison between models without/with pretrained models using \bct{} and different lifelong learning algorithms on three selected tasks from \liberolong{}.}
    \label{fig:t5-attention}
\end{figure*}
\clearpage

\textcolor{purple}{\textit{Findings:}}
Figure \ref{fig:t5-attention} shows attention map comparisons between models with/without pretrained models on \liberolong{} with \bct{} and all three LL algorithms. We observe that:
\begin{enumerate}
    \item With pretraining, the policies attend to task-irrelevant regions more easily than those without pretraining. 
    \item Some of the policies with pretraining have better attention to the task-relevant features than their counterparts without pertaining, but their performance remains lower (the last in the second row and the second in the fourth row). This observation, again, shows that there is no positive correlation between semantically meaningful attention maps and the policy's performance.
\end{enumerate}


\end{document}